\pdfoutput=1

\documentclass[11pt]{article}

\usepackage{acl}

\usepackage{times}
\usepackage{latexsym}

\usepackage[T1]{fontenc}

\usepackage[utf8]{inputenc}

\usepackage{microtype}

\usepackage{graphicx}
\usepackage{subcaption}
\newcommand{\ada}{text-ada-001}
\newcommand{\davinci}{text-davinci-002}
\usepackage{amsmath}
\usepackage{cleveref}
\usepackage{booktabs}
\usepackage{linguex}
\usepackage{makecell}
\usepackage{enumitem}
\usepackage{tikz}
\newcommand\blfootnote[1]{%
  \begingroup
  \renewcommand\thefootnote{}\footnote{#1}%
  \addtocounter{footnote}{-1}%
  \endgroup
}
\title{A fine-grained comparison of pragmatic language understanding\\in humans and language models}

\author{Jennifer Hu\textsuperscript{1}, Sammy Floyd\textsuperscript{1,2}, Olessia Jouravlev\textsuperscript{3}, Evelina Fedorenko\textsuperscript{1,4}, Edward Gibson\textsuperscript{1} \\ \\
        \textsuperscript{1}Department of Brain and Cognitive Sciences, Massachusetts Institute of Technology \\ 
        \textsuperscript{2}Department of Psychology, Sarah Lawrence College  \\
        \textsuperscript{3}Department of Cognitive Science, Carleton University \\ \textsuperscript{4}McGovern Institute for Brain Research, Massachusetts Institute of Technology\\
        \texttt{\{jennhu,samfloyd,evelina9,egibson\}@mit.edu}\\
        \texttt{olessiajouravlev@cunet.carleton.ca}
}

\begin{document}
\maketitle
\begin{abstract}
Pragmatics and non-literal language understanding are essential to human communication, and present a long-standing challenge for artificial language models. We perform a fine-grained comparison of language models and humans on seven pragmatic phenomena, using zero-shot prompting on an expert-curated set of English materials. We ask whether models (1) select pragmatic interpretations of speaker utterances, (2) make similar error patterns as humans, and (3) use similar linguistic cues as humans to solve the tasks. We find that the largest models achieve high accuracy and match human error patterns: within incorrect responses, models favor literal interpretations over heuristic-based distractors. We also find preliminary evidence that models and humans are sensitive to similar linguistic cues. Our results suggest that pragmatic behaviors can emerge in models without explicitly constructed representations of mental states. However, models tend to struggle with phenomena relying on social expectation violations.
\blfootnote{Code and data: \url{https://github.com/jennhu/lm-pragmatics}}
\end{abstract}

\section{Introduction}

Non-literal language understanding is an essential part of communication. For example, in everyday conversations, humans readily comprehend the non-literal meanings of metaphors (\emph{My new coworker is a block of ice}), polite deceits (\emph{I love the gift}), indirect requests (\emph{It's a bit cold in this room}), and irony (\emph{Classy pajamas, dude!}). These phenomena fall under the broad label of \textbf{pragmatics}, which encompasses the aspects of meaning that go beyond the literal semantics of what is said \citep{horn_semantic_1972,grice_logic_1975,yule_pragmatics_1996,levinson_presumptive_2000}. 

A long-standing challenge for NLP is to build models that capture human pragmatic behaviors. The remarkable abilities of modern language models (LMs) have triggered a recent effort to investigate whether such models capture pragmatic meaning, both through philosophical arguments \citep{bisk_experience_2020,bender_climbing_2020,potts_is_2020,michael_dissect_2020} and empirical evaluations \citep{jeretic_are_2020,zheng_grice_2021,tong_recent_2021,liu_testing_2022,ruis_large_2022,stowe_impli_2022}.
However, prior empirical studies have primarily evaluated LMs based on a binary distinction between pragmatic and non-pragmatic responses, providing limited insights into models' weaknesses. A model could fail to reach the target pragmatic interpretation in multiple ways -- for example, by preferring a literal interpretation, or by preferring a non-literal interpretation that violates certain social norms. Understanding these error patterns can suggest specific directions for improving the models, and foreshadow where pragmatics might go awry in user-facing settings \citep[e.g.,][]{saygin_pragmatics_2002,dombi_common_2022,kreiss-etal-2022-concadia}. 

From a cognitive perspective, understanding the pragmatic abilities of LMs could also offer insights into humans. Human pragmatic language comprehension involves a variety of mechanisms, such as basic language processing, knowledge of cultural and social norms \citep{trosborg_pragmatics_2010}, and reasoning about speakers' mental states \citep{brennan_two_2010,enrici_theory_2019,rubio-fernandez_pragmatic_2021}. However, it remains an open question when language understanding relies on explicit mentalizing -- which may be cognitively effortful -- versus lower-cost heuristics \citep[e.g.,][]{butterfill_how_2013,heyes_submentalizing_2014}. Because LMs lack explicit, symbolic representations of mental states, they can serve as a tool for investigating whether pragmatic competence can arise without full-blown mentalizing \citep[e.g., belief updates in the Rational Speech Act framework;][]{frank_predicting_2012}.

In this paper, we perform a fine-grained comparison of humans and LMs on pragmatic language understanding tasks. Adopting the approach of targeted linguistic evaluation \citep[e.g.,][]{linzen_assessing_2016,futrell_neural_2019,hu_systematic_2020}, our analysis serves two goals: assessing the pragmatic capabilities of modern LMs, and revealing whether pragmatic behaviors emerge without explicitly constructed mental representations. Our test materials are a set of English multiple-choice questions curated by expert researchers \citep{floyd_deciphering_nodate}, covering seven diverse pragmatic phenomena. We use zero-shot prompting to evaluate models with varying sizes and training objectives: GPT-2 \citep{radford_language_2019}, T$k$-Instruct \citep{wang_super-naturalinstructions_2022}, Flan-T5 \citep{chung_scaling_2022}, and InstructGPT \citep{ouyang_training_2022}. 

Through model analyses and human experiments, we investigate the following questions: (1) Do models recover the hypothesized pragmatic interpretation of speaker utterances? (2) When models do not select the target response, what errors do they make -- and how do these error patterns compare to those of humans? (3) Do models and humans use similar cues to arrive at pragmatic interpretations?
We find that Flan-T5 (XL) and \mbox{OpenAI's} \davinci{} achieve high accuracy and mirror the distribution of responses selected by humans. When these models are incorrect, they tend to select the incorrect literal (or straightforward) answer instead of distractors based on low-level heuristics. We also find preliminary evidence that models and humans are sensitive to similar linguistic cues. Our results suggest that some pragmatic behaviors emerge in models without explicitly constructed representations of agents' mental states. However, models perform poorly on humor, irony, and conversational maxims, suggesting a difficulty with social conventions and expectations.

\section{Related work}

Prior work has evaluated LMs' ability to recognize non-literal interpretations of linguistic input, such as scalar implicature \citep{jeretic_are_2020,schuster_harnessing_2020,li_predicting_2021} or figurative language \citep{tong_recent_2021,liu_testing_2022,gu_just-dream-about-it_2022,stowe_impli_2022}. In a broad-scale evaluation, \citet{zheng_grice_2021} test five types of implicatures arising from \citeauthor{grice_logic_1975}'s (\citeyear{grice_logic_1975}) conversational maxims, and evaluate their models after training on the task. In our work, we consider Gricean implicatures as one of seven phenomena, and we evaluate pre-trained LMs without fine-tuning on our tasks.

Similar to our work, \citet{ruis_large_2022} also use prompting to evaluate LMs on pragmatic interpretation tasks. They formulate implicature tests as sentences ending with ``yes'' or ``no'' (e.g., ``Esther asked “Can you come to my party on Friday?” and Juan responded “I have to work”, which means no.''). A model is considered pragmatic if it assigns higher probability to the token that makes the sentence consistent with an implicature. In our work, models must select from multiple interpretations, enabling a detailed error analysis and comparison to humans. \citeauthor{ruis_large_2022}'s materials also focus on indirect question answering as an implicature trigger, whereas we consider a broader range of pragmatic phenomena and utterance types.

Since pragmatic language understanding often draws upon knowledge of social relations, our tasks are conceptually related to benchmarks for evaluating social commonsense \citep[e.g.,][]{sap_social_2019,zadeh_social-iq_2019}. These evaluations focus on the interpretation of actions and events, whereas we focus on the interpretation of speaker utterances. Another hypothesized component of pragmatics is Theory of Mind \citep[ToM;][]{leslie_core_2004,apperly_mindreaders_2011}, or the ability to reason about others' mental states. Benchmarks for evaluating ToM in models \citep[e.g.,][]{nematzadeh_evaluating_2018,le_revisiting_2019,sap_neural_2022} primarily focus on false-belief tasks \citep{baron-cohen_does_1985}, which assess whether a model can represent the beliefs of another agent that are factually incorrect but consistent with that agent's observations. LMs have been shown to succeed on some ToM tests \citep{kosinski_theory_2023} while failing on others \citep{sap_neural_2022,ullman_large_2023}. 

\section{Evaluation materials} \label{sec:tasks}

\subsection{Overview of stimuli} \label{sec:stimuli-overview}

\renewcommand{\cellalign}{tl}
\renewcommand{\theadalign}{tl}

\newcommand{\blob}[2]{
    \vcenter{\hbox{\begin{tikzpicture}
    \node[rounded corners,fill=#1] {\sffamily\tiny{\color{white}{#2}}};
    \end{tikzpicture}}}
}
\definecolor{correct}{rgb}{0.55, 0.71, 0.0}
\definecolor{incorrectnonliteral}{rgb}{0.8, 0.25, 0.33}
\definecolor{incorrectliteral}{rgb}{0.29, 0.59, 0.82}

\newcommand{\literal}{${\blob{incorrectliteral}{Literal}}$}
\newcommand{\correctnonliteral}{${\blob{correct}{{Correct}}}$}
\newcommand{\nonliteral}[1]{$\blob{incorrectnonliteral}{#1}$}
\begin{table*}[ht]

    \centering
    \scriptsize
    \begin{tabular}{p{1cm}p{3.8cm}p{9.3cm}} \toprule
        Task & Example query & Example answer options \\ \midrule
        Deceits & Henry is sitting at his desk and watching TV, and reluctantly switches off the TV with the remote control and picks up a textbook. Shortly after, his mother comes in the room and asks, "What have you been doing up here?" Henry responds: "Reading." Why has Henry responded in such a way? & \begin{enumerate}[topsep=0ex,itemsep=-1ex,partopsep=0ex,parsep=1ex]
            \item \correctnonliteral{} He does not want to get into trouble for not studying.
            \item \literal{} He has been reading for some time.
            \item \nonliteral{DistractorLexicalOverlap} He does not want to offend his mom by not reading the books that she gave him.
            \item \nonliteral{DistractorSocialConvention} He wants his mom to believe that he has been watching TV.
            \end{enumerate} \\ \midrule
        Indirect speech & Nate is about to leave the house. His wife points at a full bag of garbage and asks: "Are you going out?" What might she be trying to convey? & \begin{enumerate}[topsep=0pt,itemsep=-1ex,partopsep=0ex,parsep=1ex]
            \item \correctnonliteral{} She wants Nate to take the garbage out. 
            \item \literal{} She wants to know Nate's plans.
            \item \nonliteral{DistractorAssociative} She wants Nate to bring his friends over.
            \item \nonliteral{DistractorLexicalOverlap} She wants Nate to spend more time with the family.
            \end{enumerate}\\ \midrule
        Irony & It is a holiday. Stefan and Kim are sitting in the backseat of the car. They are fighting all the time. Their father says: "Oh, it is so pleasant here." What did the father want to convey? & \begin{enumerate}[topsep=0pt,itemsep=-1ex,partopsep=0ex,parsep=1ex]
            \item \correctnonliteral{} He does not want to listen to his kids' arguments.
            \item \literal{} He enjoys listening to his kids fighting. 
            \item \nonliteral{DistractorAssociative} AC gives them some needed cool.
            \item \nonliteral{DistractorNonSequitur} He remembers about his wife's birthday.
            \end{enumerate} \\ \midrule
        Maxims & Leslie and Jane are chatting at a coffee shop. Leslie asks, "Who was that man that I saw you with last night?" Jane responds, "The latte is unbelievable here." Why has Jane responded like this? & 
        \begin{enumerate}[topsep=0pt,itemsep=-1ex,partopsep=0ex,parsep=1ex]            
            \item \correctnonliteral{} She does not want to discuss the topic that Leslie has raised.
            \item \literal{} She thinks that it is the best latte in the town.
            \item \nonliteral{DistractorAssociative} The man who Leslie saw makes unbelievable lattes.
            \item \nonliteral{DistractorNonLiteral} A coffee break is not a good time to discuss men.
        \end{enumerate} \\ \midrule
        Metaphor & Andrew and Bob were discussing the investment company where Andrew works. Bob said: ``The investors are squirrels collecting nuts.'' What does Bob mean? & \begin{enumerate}[topsep=0pt,itemsep=-1ex,partopsep=0ex,parsep=1ex]
        \item \correctnonliteral{} They buy stocks hoping for future profit.  
        \item \literal{} Squirrels were hired to work in the company. 
        \item \nonliteral{DistractorNonLiteral} The investors dress and eat well.
        \item \nonliteral{DistractorNonSequitur} Bob is allergic to nuts.
        \item \nonliteral{DistractorPlausibleLiteral} The investors enjoy picking nuts as much as squirrels do.
        \end{enumerate} \\ \midrule
        Humor & Martha walked into a pastry shop. After surveying all the pastries, she decided on a chocolate pie. "I'll take that one," Martha said to the attendant, "the whole thing." "Shall I cut it into four or eight pieces?" the attendant asked. & 
        \begin{enumerate}[topsep=0pt,itemsep=-1ex,partopsep=0ex,parsep=1ex]
            \item \correctnonliteral{} Martha said, "Four pieces, please; I'm on a diet." 
            \item $\blob{incorrectliteral}{Literal}$ Martha said: "Well, there are five people for dessert tonight, so eight pieces will be about right."
            \item \nonliteral{DistractorAssociative} Martha said, "You make the most delicious sweet rolls in town."
            \item \nonliteral{DistractorFunny} Then the attendant squirted whipped cream in Martha's face.
            \item \nonliteral{DistractorNeutral} Martha said, "My leg is hurting so much."
        \end{enumerate} \\ \midrule
        Coherence & Mary's exam was about to start. Her palms were sweaty. & 
        \begin{enumerate}[topsep=0pt,itemsep=-1ex,partopsep=0ex,parsep=1ex]
            \item $\blob{correct}{Correct}$ Coherent
            \item $\blob{incorrectnonliteral}{Incorrect}$ Incoherent
        \end{enumerate}
        \\\bottomrule
            
    \end{tabular}
    \caption{Sample item from each task in our evaluation. All items are originally curated by \citet{floyd_deciphering_nodate}.}
    \label{tab:examples}
\end{table*}

Our evaluation materials are taken from \citeauthor{floyd_deciphering_nodate}'s (\citeyear{floyd_deciphering_nodate}) experiments,\footnote{Materials can be found at \url{https://osf.io/6abgk/?view_only=42d448e3d0b14ecf8b87908b3a618672}.} covering seven phenomena. Each item is a multiple choice question, with answer options representing different types of interpretation strategies. For most of the tasks, the question has three parts: a short story context (1-3 sentences), an utterance by one of the characters, and a question about what the character intended to convey.\footnote{The exceptions are Humor and Coherence.} \Cref{tab:examples} shows an example item for each task, with annotated answer options. $\blob{correct}{Green}$ labels indicate the target pragmatic interpretation.\footnote{We refer to these answer options as ``Correct'' throughout the paper. However, these answers are only ``correct'' in the sense of a normative evaluation. We acknowledge the wide variation in individual humans' abilities and tendencies to use non-literal language, which is not captured in our analyses. We thank an anonymous reviewer for highlighting this point.} $\blob{incorrectliteral}{Blue}$ labels indicate the literal interpretation. $\blob{incorrectnonliteral}{Red}$ labels indicate incorrect non-literal interpretations, which are based on heuristics such as lexical similarity to the story, thus serving as distractor options.

Each task has 20-40 items, which were manually curated by expert researchers to cover a broad range of non-literal phenomena and elicit individual differences among humans. The stimuli were not specifically designed to require Theory of Mind reasoning (ToM). However, behavioral and neural evidence suggests that many of the tested phenomena rely on mentalizing processes. In \Cref{sec:tested-phenomena}, we briefly describe the role of ToM for each tested phenomenon, and how LMs' training corpora may provide linguistic cues to perform the tasks.

\subsection{Tested phenomena} \label{sec:tested-phenomena}

\paragraph{Deceits.} Humans produce polite deceits (``white lies'') in the service of social and personal relationships \citep[e.g.,][]{camden_white_1984}. Behavioral studies in young children suggest that understanding white lies requires interpretive ToM, or the ability to allow different minds to interpret the same information in different ways \citep{hsu_two_2013}. Furthermore, the tendency to produce white lies is linked to emotional understanding abilities, \citep{demedardi_prosocial_2021}, and moral judgments about white lies are linked to second-order false-belief understanding \citep{vendetti_theory_2019}.

The Deceits task presents a story with a white lie, and asks why the speaker has used this utterance. The underlying intentions behind polite deceits are rarely explicitly explained in text. As a result, it is unlikely that LMs learn a direct connection between the utterance and the speaker's intention during training on static texts. However, instances of polite deceits in text corpora may be accompanied by descriptions of characters' emotional states, which may indicate that speakers' intentions differ from what is literally conveyed by their utterance. This highlights the importance of context in interpreting deceits, which we return to in \Cref{sec:role-of-context}. 

\paragraph{Indirect speech.} Humans often use language in a performative sense, such as indirectly requesting an action from other individuals  \citep[e.g.,][]{austin_how_1975,searle_indirect_1975}. Indirect or polite speech comprehension has been captured by Rational Speech Act \citep[RSA;][]{frank_predicting_2012} models, which characterize listeners as performing Bayesian inference about a speaker who chooses utterances based on a tradeoff between epistemic and social utility \citep{brown_politeness_1987,yoon_talking_2016,yoon_polite_2020,lumer_modeling_2022}.

The IndirectSpeech task presents a story with an indirect request, and asks what the speaker intends to convey. Like deceits, it's unlikely that indirect speech acts are explained in text data. However, indirect requests may be followed by descriptions of the completion of the implied request -- for example, that someone closed a window after hearing the utterance ``It's cold in here''. Therefore, models may learn relationships between the utterances and desired outcomes through linguistic experience.

\paragraph{Irony.} Humans use irony to convey the opposite of the semantic content of their utterance  \citep{booth_rhetoric_1974,wilson_verbal_1992,attardo_irony_2000,wilson_meaning_2012}. As such, irony has long been hypothesized to rely on social reasoning and perspective-taking \citep[e.g.,][]{happe_communicative_1993,andres-roqueta_contribution_2017}. Indeed, human irony comprehension behaviors are captured by Bayesian reasoning models that take into account speakers' affective goals \citep{kao_lets_2014}. In addition, neuroimaging studies suggest that irony interpretation relies on brain regions that are implicated in classic ToM tasks \citep{spotorno_neural_2012}.

The Irony task presents a story with an ironic statement, and asks what the character intends to convey. While ironic statements are also rarely explained in text, models could leverage accompanying cues such as descriptions of characters' emotional states or a mismatch in sentiment. 

\paragraph{Maxims of conversation.} \citet{grice_logic_1975} proposes that communication follows a set of \emph{maxims}: be truthful; be relevant; be clear, brief, and orderly; and say as much as needed, and no more. A prevailing theory is that listeners derive implicatures by expecting speakers to be cooperative (i.e., abide by the maxims) and reasoning about speakers' beliefs and goals. Indeed, there is extensive evidence for RSA models capturing these implicatures, such as those arising from the maxims of \emph{quantity} \citep{potts_embedded_2016,frank_rational_2018,degen_rational_2023} and \emph{manner} \citep{bergen_pragmatic_2016,franke_probabilistic_2016,tessler_not_2018}. 

The Maxims task presents a story with a character flouting one of Grice's maxims, and asks why the character has responded in such a way. Based on linguistic input, it may be easy for LMs to recognize when a speaker is flouting a maxim -- for example, if an utterance is particularly long, features an uncommon syntactic construction, or diverges semantically from the context. However, it is unclear whether LMs will be able to recover the speaker's underlying intentions.

\paragraph{Metaphor.} Metaphors \citep{lakoff_metaphors_1980} are used to draw comparisons between entities in a non-literal sense. Metaphor understanding has been hypothesized to require mentalizing \citep{happe_communicative_1993}, and fine-grained metaphor comprehension behaviors are captured by RSA models where listeners and speakers reason about each others' beliefs and goals \citep{kao_formalizing_2014}.

The Metaphor task presents a story with a metaphor, and asks what the speaker intends to convey. For models, the challenges of metaphor comprehension include accessing world knowledge and forming abstract relationships between domains. However, it is possible that the relevant properties of the entities under comparison could emerge through linguistic experience. 

\paragraph{Humor.} Humor is one of the most distinctive aspects of human conversation, reflecting communicative goals with complex social function \citep{veatch_theory_1998,martin_psychology_2018}. Neuroimaging studies suggest that joke understanding is supported by regions in the ToM brain network \citep{kline_struhl_understanding_2018}. Behavioral tests also reveal associations between ToM and humor abilities \citep{aykan_assessing_2018,bischetti_pragmatics_2019}.

The Humor task presents a joke and asks which punchline makes the joke the funniest.\footnote{Unlike the other tasks, there is no speaker utterance.} Some theories argue that humor is triggered by linguistic incongruency effects \citep[e.g.,][]{deckers_humor_1975}, which might be straightforward for LMs to detect. Recent work has also shown that LMs can explain certain jokes \citep{chowdhery_palm_2022}. However, some of \citeauthor{floyd_deciphering_nodate}'s Humor items require complex world knowledge -- for example, that slicing a pie into four versus eight pieces does not change the total amount of pie (see \Cref{tab:examples}). As such, selecting the funniest punchline is a nontrivial task.

\paragraph{Coherence inferences.} Humans also make pragmatic inferences beyond the sentence level -- for example, by assuming that consecutive sentences form a logical or sequential relationship. \citet{moss_comprehension_2015} and \citet{jacoby_discourse-level_2020} find that constructing these discourse relationships loads on regions of the ToM brain network, suggesting a role of ToM in coherence inferences.

The Coherence task presents a pair of sentences, and asks whether the pair forms a coherent story.\footnote{This task differs from the others in that there is no speaker utterance, and the answer options are identical across items (``Coherent'' or ``Incoherent'').} We assume that LMs' training data, which consists of naturalistic text, is primarily coherent. Therefore, we expect LMs to be able to distinguish between coherent and incoherent sentence pairs \citep[for an in-depth study, see][]{beyer_is_2021}.

\section{Experiments}

\subsection{Evaluation paradigm} \label{sec:evaluation}
Our evaluation paradigm uses \emph{zero-shot prompting}. Prompting can easily be adapted to all of our seven tasks, allowing us to compare performance across tasks within a model.  Prompting also allows us to present models with inputs that are nearly identical to the stimuli seen by humans in \citeauthor{floyd_deciphering_nodate}'s experiments, whereas other methods would require converting the stimuli into task-specific formats. We choose zero-shot prompts in order to evaluate the knowledge that emerges through training, and not through in-context adaptation to the task.

\paragraph{Prompt structure.} Each prompt consists of two parts: task instructions, and a query. The instructions are nearly identical to the instructions presented to humans in \citeauthor{floyd_deciphering_nodate}'s experiments, prepended with the keyword ``Task:''. The only other modification is that the original instructions had a final sentence of ``Please answer as quickly as possible'', which we replaced with a sentence like ``The answer options are 1, 2, 3, or 4''.\footnote{The exact answer options changed according to the task.} 

For all tasks except Humor, the query consists of the scenario (prepended with keyword ``Scenario:'') and question, and then the numbered answer options (prepended with ``Options:'').\footnote{For the Humor task, the joke is prepended with ``Joke:'', and the answer options are prepended with ``Punchlines:''.} The prompt concludes with the keyword ``Answer:''. Full example prompts are given in \Cref{sec:example-prompts}.

\paragraph{Evaluation.} To evaluate a model on a given item, we feed the prompt to the model, and measure the model's probability distribution over tokens conditioned on the prompt. We compare the probabilities of each answer token (e.g., ``1'', ``2'', ``3'', or ``4'') under this distribution. The model is considered correct on a given item if it assigns highest probability to the correct answer token, among all the possible answer tokens for that item. 

We generated 5 versions of each item by randomizing the order of answer options. This was done to control for the base probabilities of the answer tokens. Since we do not analyze generated text, the model results themselves are deterministic.

\begin{table}[t]
    \scriptsize
    \centering
    \begin{tabular}{lrc} \toprule
        Model & \# parameters & Training \\ \midrule
        GPT-2 & 117M & Autoregressive LM \\
        Tk-Instruct (3B) & 3B & Multitask \\
        Tk-Instruct (11B) & 11B & Multitask \\
        Flan-T5 (base) & 250M & Multitask \\
        Flan-T5 (XL) & 3B & Multitask \\
        InstructGPT-3 (ada) & 350M (est.) & Multitask, human feedback \\
        \davinci & Unknown & FeedME \\\bottomrule
    \end{tabular}
    \caption{Models tested in our experiments.}
    \label{tab:models}
\end{table}

\begin{figure*}[t]
    \centering
    \includegraphics[width=\linewidth]{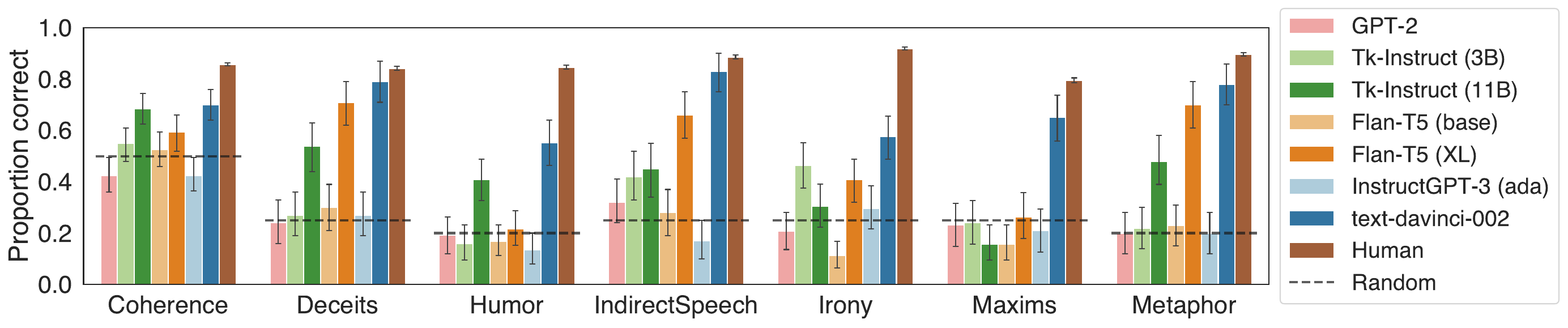}
    \caption{Accuracy for each task. Error bars denote 95\% CI. Dashed line indicates task-specific random baseline.}
    \label{fig:accuracy}
\end{figure*}

\begin{figure}[t]
    \centering
    \includegraphics[width=0.95\linewidth]{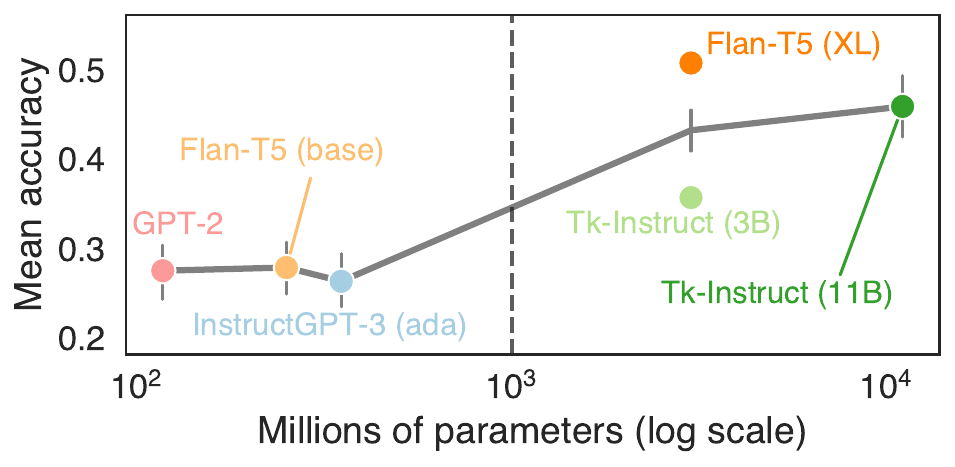}
    \caption{Mean accuracy vs.~millions of parameters. Vertical dashed line indicates 1 billion parameters. \davinci{} was excluded from this analysis, as the number of parameters is unknown.}
    \label{fig:accuracy_vs_size}
\end{figure}

\subsection{Models}

We test seven models across four model families, summarized in \Cref{tab:models}.\footnote{All non-OpenAI models were accessed via Huggingface \citep{wolf_transformers_2020} and run on a single NVIDIA A100 GPU.} As a baseline, we first test a base \textbf{GPT-2} model \citep[117M parameters;][]{radford_language_2019}, which is trained on an autoregressive language modeling objective. 

Second, we test a set of models which are based on T5 \citep{raffel_exploring_2020} and instruction-finetuned on a diverse collection of tasks \citep{wei_finetuned_2022}. This set of models consists of two \textbf{T$k$-Instruct} models \citep[3B and 11B;][]{wang_super-naturalinstructions_2022}, which were fine-tuned on 1.6K tasks, and two \textbf{Flan-T5} models \citep[base: 250M parameters; XL: 3B parameters;][]{chung_scaling_2022}, which were fine-tuned on 1.8K tasks. The fine-tuning tasks cover a wide range of categories, such as commonsense reasoning, translation, mathematics, and programming.

Finally, we test two \textbf{InstructGPT}-based models \citep{ouyang_training_2022} via the OpenAI API: \ada{} (350M parameters), which we refer to as InstructGPT-3 (ada); and \davinci{}, which comes from the GPT-3.5 family of models.\footnote{Parameter estimates come from \url{https://blog.eleuther.ai/gpt3-model-sizes/}. Although the size of \davinci{} is unknown, we assume that it is larger than InstructGPT-3 (ada).}\textsuperscript{,}\footnote{The OpenAI model results might not be reproducible, but timestamps of API calls can be found in \Cref{sec:openai-timestamps}.} These models are fine-tuned to follow instructions and align with human feedback.

We compare models to a baseline from 374 humans, collected by \citet{floyd_deciphering_nodate}. Their experiments presented multiple choice questions to humans in nearly identical format to our prompts.

\section{Results}

We now return to the three questions posed in the Introduction, in each of the following subsections. 

\begin{figure*}[ht]
    \centering
    \includegraphics[width=\linewidth]{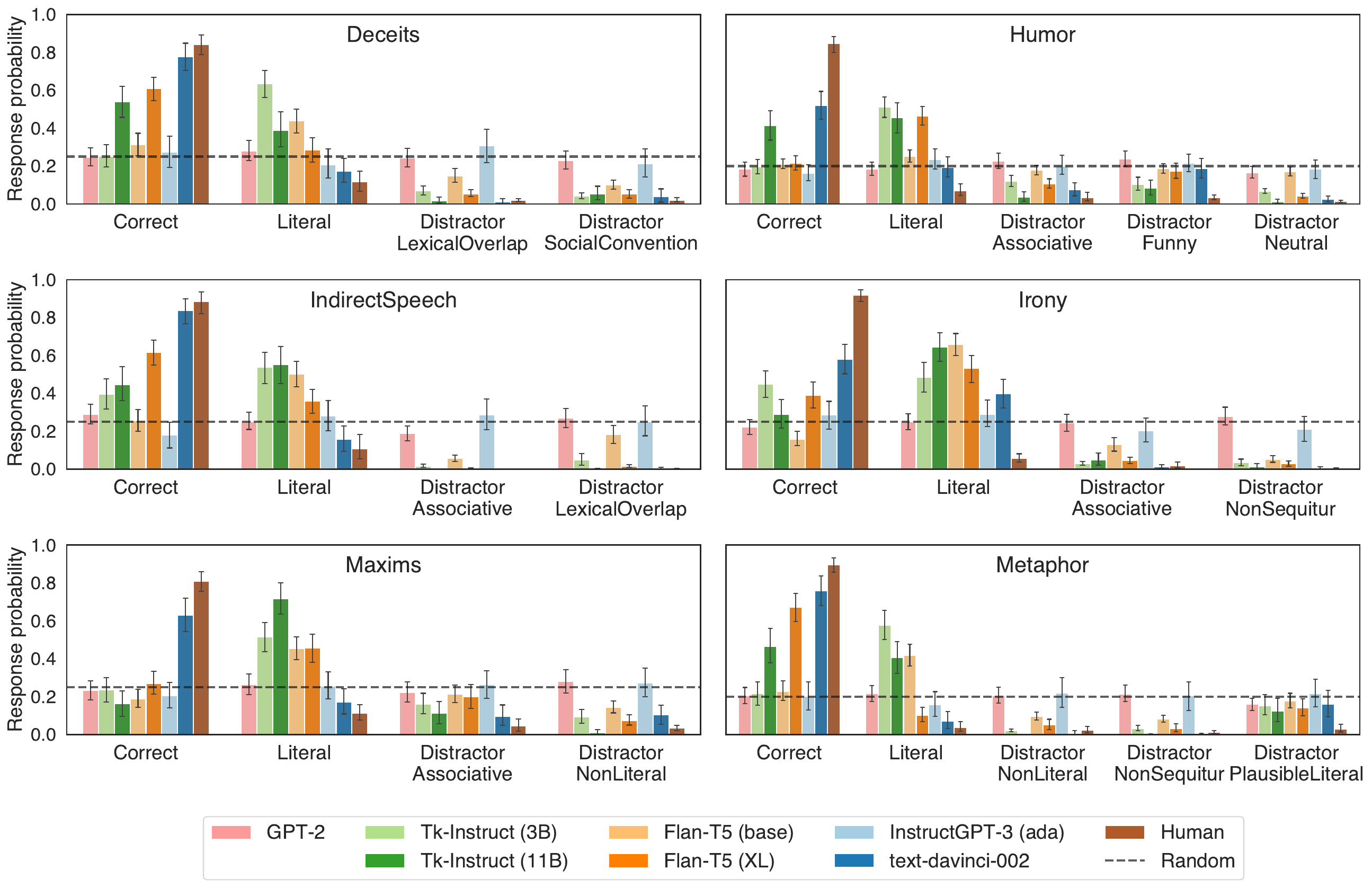}
    \caption{Response distributions across models and humans. Answer options for each task are shown on the x-axis. For models, y-axis denotes probability assigned to each answer option. For humans, y-axis denotes empirical frequency of each answer option being selected. Error bars denote 95\% CI. Dashed line indicates random baseline.}
    \label{fig:answer-dists}
\end{figure*}

\subsection{Do models choose the target pragmatic interpretation?}

\Cref{fig:accuracy} shows the proportion of trials where models and humans select the pragmatic answer. The smallest models (GPT-2, Flan-T5 (base), InstructGPT-3 (ada)) fail to perform above chance. The largest models (T$k$-Instruct (11B), Flan-T5 (XL), \davinci) perform above chance on all tasks (except T$k$-Instruct (11B) on Maxims), and in some cases near human-level. Overall, models perform worst at the Humor, Irony, and Maxims tasks. Interestingly, these phenomena involve speakers violating listeners' expectations in some way: producing a funny punchline to a mundane story (Humor), stating the direct opposite of the speaker's belief (Irony), or disobeying one of the assumed rules of conversation (Maxims). It may be that models fail to represent certain social expectations that are maintained by human listeners.

Next, we investigated the relationship between model size and accuracy. \Cref{fig:accuracy_vs_size} shows the mean accuracy achieved by each model (averaged across tasks) vs.~millions of parameters. The line and error bars denote the mean and 95\% CIs, while points represent individual models. We find a coarse effect of model size: there is a stark jump in accuracy after 1B parameters (dashed line). However, model size does not fully explain variance in accuracy: all models with $<$1B parameters achieve similar accuracy, and Flan-T5 (XL) outperforms T$k$-Instruct (3B), despite both having 3B parameters. 

\subsection{Do models and humans make similar types of errors?}

Recall from \Cref{sec:tasks} that each item has a set of answer options that correspond to different strategies (\Cref{tab:examples}).\footnote{The exception is Coherence, which is excluded here.} In addition to the target pragmatic answer (Correct), each item also has a plausible but unlikely literal answer (Literal), as well as distractors based on lexical overlap or semantic associations (Distractor*). For each item, we computed the human empirical distribution over answer choices, and compared it to models' probability assigned to the answer tokens (e.g., ``1'', ``2'', ``3'', and ``4''). 

\Cref{fig:answer-dists} shows the answer distributions for each task. Across tasks, humans primarily select the Correct option, occasionally the Literal option, and rarely the distractors. We find a similar pattern for \davinci{}, although the model is more likely to select the Literal option in general. The other large models (T$k$-Instruct (11B), Flan-T5 (XL)) also generally assign highest probability to the Correct and Literal options, although the distribution looks less human-like. The next-largest models (T$k$-Instruct (3B), Flan-T5 (base)) prefer the Literal option, and the remaining models (GPT-2, InstructGPT-3 (ada)) are at chance. These results show that larger models consistently identify the literal interpretation of an utterance, suggesting that their pragmatic failures are unlikely to be explained by a failure to represent basic semantic meaning (for our test materials). 

However, even high-performing models occasionally do select the distractor answers, revealing interesting behaviors. For example, in the Metaphor task, \davinci{} and Flan-T5 (XL) prefer the DistractorPlausibleLiteral option -- which is a figurative reading of the utterance -- over the Literal option -- which is completely non-figurative. Similarly, in the Humor task, \davinci{} is much more likely to select the DistractorFunny option over the other (non-humorous) distractors. This suggests a coarse sensitivity to humor, even if the model selects the human-preferred punchline only 55\% of the time (see \Cref{fig:accuracy}). We take this analysis to illustrate the value of looking beyond binary pragmatic/non-pragmatic response distinctions, and using controlled distractor items to evaluate models' abilities \citep[e.g.,][]{mccoy_right_2019}.

\subsection{Are models and humans sensitive to similar linguistic cues?}

Having found qualitatively similar response patterns between humans and models, we now ask \emph{how} models and humans arrive at pragmatic interpretations, and whether they use similar types of information. We begin with a broad evaluation of the extent to which models and humans rely on linguistic context (\Cref{sec:role-of-context}). We then take a more granular approach and ask whether model and human performance is correlated at the item level -- i.e., if models and humans exhibit similar sensitivity to the cues that make a non-literal interpretation more or less likely (\Cref{sec:by-item-difficulty}).

\subsubsection{The role of context} \label{sec:role-of-context}

Many cues for enriched language understanding come from the context in which the speaker makes their utterance.
However, some aspects of non-literal comprehension might arise given the utterance in isolation, while others are highly sensitive to specific contextual details \citep[e.g.,][]{levinson_presumptive_2000}. Therefore, we expect that the degree to which humans rely on context to select non-literal interpretations will vary across the tested tasks. 

To investigate this variation, we created a new set of stimuli by removing the context stories, leaving only the speaker utterance and final question (e.g., \emph{Dan says, ``The dog knocked it over.'' Why has Dan responded in such a way?}).\footnote{This manipulation is not compatible with the Humor and Coherence tasks, so they are excluded from this analysis.} 
We re-ran the human experiment on 30 participants, following the protocols of \citet{floyd_deciphering_nodate}'s original experiment using the no-context modified materials.\footnote{Details can be found in \Cref{sec:human-expts}.} We also re-ran the three models that achieved highest accuracy on the original items: T$k$-Instruct (11B), Flan-T5 (XL), and \davinci{}.

\begin{figure}[t]
    \centering
    \includegraphics[width=0.95\linewidth]{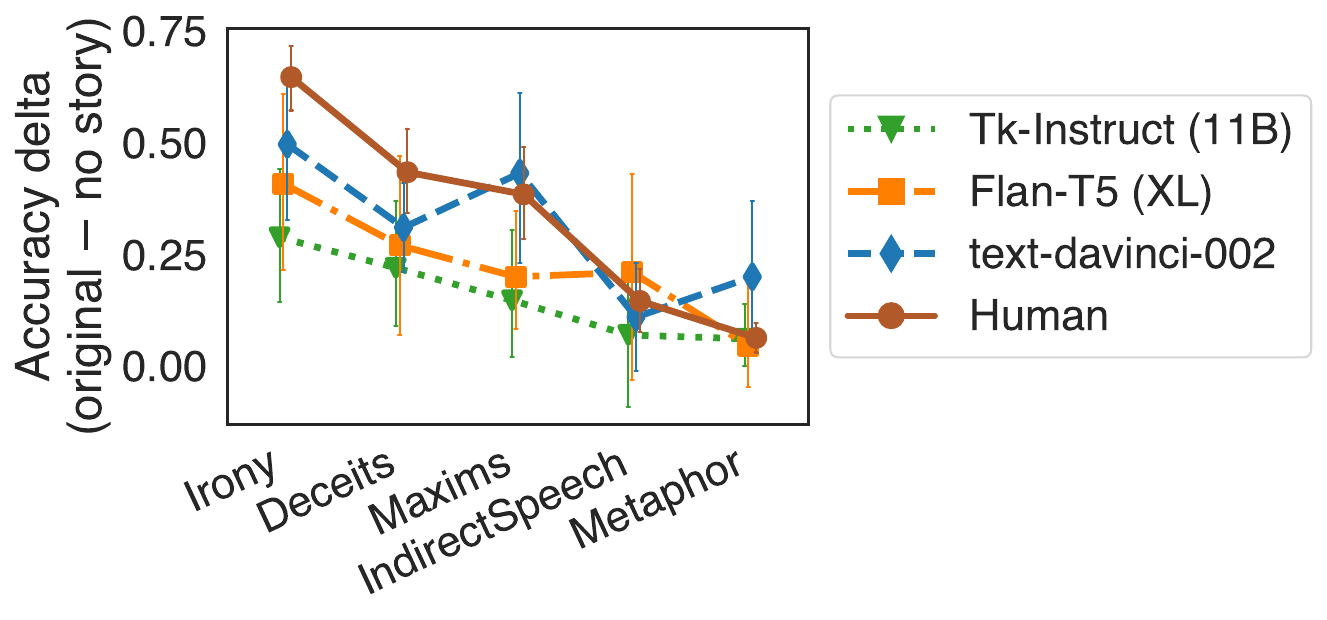}
    \caption{Mean by-item difference in accuracy once story context was removed.}
    \label{fig:acc-no-story}
\end{figure}

\Cref{fig:acc-no-story} shows the mean accuracy difference on the original versus no-context versions of each item.\footnote{See \Cref{fig:no-context-raw} in \Cref{sec:no-context-raw} for comparison of raw accuracy scores on the original and no-context items.} We find that models and humans exhibit a similar qualitative pattern: removing the story leads to the largest degradation for Irony, followed by Deceits and Maxims. This aligns with our intuitions, because in these cases, speakers' utterances can be interpreted either literally or as the complete opposite, based on the specific social situation (e.g., ``It is so pleasant here''). In contrast, there are smaller degradations for IndirectSpeech and Metaphor. This suggests that some indirect requests are conventionalized (e.g., ``I am getting cold''), although their interpretations may be facilitated by context \citep[e.g.,][]{gibbs_contextual_1979}. Similarly, this suggests that metaphor interpretation may draw more upon global knowledge than local context.

\subsubsection{Scrambling} 

Next, we tested whether models rely on syntactic and discourse-level information from the context, or whether they can perform the tasks when ordering cues are removed. We constructed two scrambled versions of each item by randomizing the order of sentences and words. In both versions, the instructions, final question (e.g., \emph{Why has Dan responded in such a way?}), and answer options were unmodified and remained in their original positions. Again, we only tested the best-performing models on these items.

We found that models maintain reasonable performance for most tasks, with the notable exception of Metaphor (\Cref{fig:scrambling}; \Cref{sec:scrambling}). This robustness to scrambling accords with prior evidence that models often rely on lexical information without human-like compositionality \citep[e.g.,][]{dasgupta_evaluating_2018,nie_analyzing_2019,mccoy_right_2019}. We expect that scrambling, especially at the word-level, would likely disrupt human performance, but this remains an open empirical question. We leave an investigation of human performance to future work.

\subsubsection{Item-level alignment} \label{sec:by-item-difficulty}

Up to this point, we analyzed differences across phenomena by averaging over items. However, there is also variance \emph{within} each phenomenon in the types of cues that suggest how the utterances should be interpreted. For example, some items contain explicit descriptions of characters' emotional states (e.g., ``Sarah becomes angry''). If models and humans leverage these cues in similar ways, then we would expect to see correlations between model and human performance at the item level.

\begin{figure}[t]
    \centering
    \includegraphics[width=0.85\linewidth]{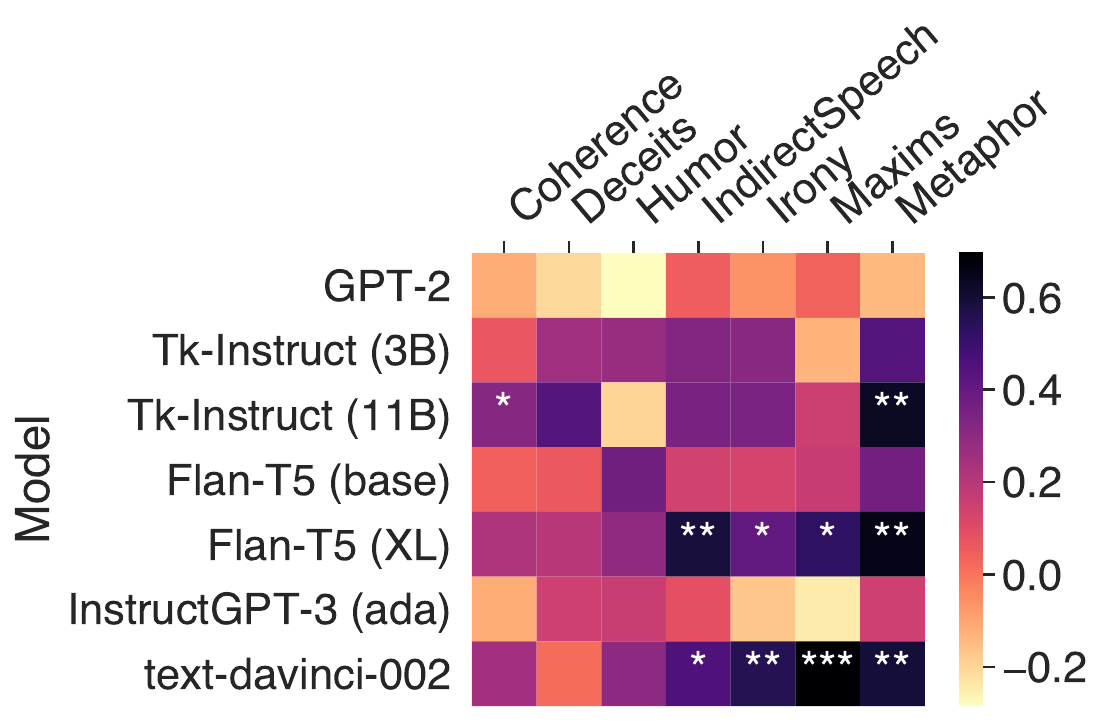}
    \caption{Pearson correlation coefficients between by-item human accuracy and model probability of the correct answer. Cells are marked with significance codes.}
    \label{fig:by-item-corr}
\end{figure}

For each task and model, we compute the Pearson correlation between by-item mean accuracy achieved by humans and by-item mean probability that models assigned to the correct answer (\Cref{fig:by-item-corr}). In general, the larger models (T$k$-Instruct (11B), Flan-T5 (XL), \davinci) are better aligned with humans, and the strongest correlations occur for IndirectSpeech, Irony, Maxims, and Metaphor. 
This suggests that for those tasks, models and humans are similarly sensitive to cues  that make a non-literal interpretation likely.

\section{Discussion}

We used an expert-curated set of materials \citep{floyd_deciphering_nodate} to compare LMs and humans on seven pragmatic phenomena. We found that Flan-T5 (XL) and \davinci{} achieve high accuracy and match human error patterns: within incorrect responses, these models tend to select the literal interpretation of an utterance over heuristic-based distractors. We also found preliminary evidence that LMs and humans are sensitive to similar linguistic cues: model and human accuracy scores correlate at the item level for several tasks, and degrade in similar ways when context is removed. 

Our results suggest that language models can consistently select the pragmatic interpretation of a speaker's utterance -- but how? The models tested in our experiments reflect a variety of learning processes through which pragmatic knowledge could emerge. GPT-2 is trained to learn the distribution of linguistic forms; the T$k$-Instruct and Flan-T5 models are pre-trained on a denoising task and fine-tuned on thousands of instruction-based tasks; and the OpenAI models receive signal from human feedback. Our experiments are not designed to tease apart the contributions of these training procedures to models' behaviors. Therefore, we do not intend to make strong claims about the mechanisms by which models learn pragmatics. 

A shared feature of our tested models is the lack of explicitly constructed mental state representations. In this sense, our results are potentially compatible with two hypotheses. One possibility is that the models do not have an ability that can be considered an analog of Theory of Mind (ToM). This view is supported by evidence that language models perform poorly on social commonsense and false-belief tasks \citep{sap_neural_2022}, and are remarkably brittle to small perturbations of classic tests \citep{ullman_large_2023}. If models truly lack ToM, then their pragmatic behaviors might be explained by inferences based on low-level linguistic cues. Taken a step further, this finding could potentially suggest that certain human pragmatic behaviors arise through inferences based on language statistics, with no need for mental state representations.

A second possibility is that models do have a heuristic version of ToM, which is not explicitly engineered but instead emerges as a by-product of optimizing for other objectives (such as linguistic prediction). Since language contains many descriptions of agents' beliefs, emotions, and desires, it may be beneficial -- perhaps even necessary -- to induce representations of these mental states in order to learn a generative model of linguistic forms. Indeed, \citet{andreas_language_2022} argues that whereas language models have no explicit representation of communicative intents, they can infer approximate representations of the mental states of the agents that produce a given linguistic context. If this hypothesis is true, however, it would still remain unclear whether ToM is \emph{necessary} to support the pragmatic behaviors tested in our evaluation materials.

Our experiments do not differentiate between these two hypotheses. However, fine-grained behavioral evaluations -- such as those presented in this work -- are important for revealing models' capabilities and weaknesses, and offer a first step toward understanding how pragmatic behaviors can be supported. A promising direction for future work is to test models with a wider range of training objectives, or even new architectures, such as distinct language and social reasoning modules \citep[see][]{mahowald_dissociating_2023}. 
In addition, although there is evidence for the role of mentalizing in our tested pragmatic phenomena (see \Cref{sec:stimuli-overview}), one limitation of our stimuli is that they were not specifically designed to require ToM. New datasets that perform targeted manipulations of ToM alongside tests of language comprehension could help reveal how linguistic experience and ToM jointly support pragmatic behaviors. 

\section*{Acknowledgments}

We would like to thank the anonymous reviewers as well as Roger Levy, Christopher Potts, and Josh Tenenbaum for their constructive feedback. We also thank Quinn Langford for help with coding details of the stimuli. This work was in part supported by a grant from the Simons Foundation to the Simons Center for the Social Brain at MIT. J.H.~is supported by an NSF Graduate Research Fellowship (\#1745302) and an NSF Doctoral Dissertation Research Improvement Grant (BCS-2116918). S.F.~is funded by the NSF SPRF (\#2105136). E.F.~was additionally supported by NIH award R01-DC016950 and by research funds from the McGovern Institute for Brain Research and the Department of Brain and Cognitive Sciences.

\section*{Limitations}

We note several methodological limitations with our experiments. First, since the evaluation materials were manually crafted, there is a rather small number of items (compared to the size of automatically generated NLP benchmarks). Small evaluation sets can introduce issues of statistical power \citep{card_little_2020} and introduce bias based on lexical items. We feel this is not a major concern, because (1) our materials are validated by expert researchers; (2) models can be directly compared to humans in \citeauthor{floyd_deciphering_nodate}'s experiments; and (3) in practice, there is enough signal to distinguish between the tested models. 

Second, we only evaluate models on English-language materials, and some of the tasks were designed based on norms of communication and social interaction in Western cultures. As pragmatics can vary widely across language and cultures \citep{li_cultural_2012,rubio-fernandez_incrementality_2020,floyd_conversation_2021,brown_cross-speaker_2021,dideriksen_language_2022}, an important direction for future work is to evaluate pragmatics beyond English \citep{ameka_what_2019,blasi_over-reliance_2022}. 

Third, aside from the OpenAI API models, we were only able to test models with $\leq$11B parameters due to limited computational resources. Models with parameter sizes between 11B and the size of \davinci{} could exhibit qualitatively different behaviors.

Finally, we emphasize that it is impossible to predict how models will respond to an arbitrary input. Therefore, we caution against extrapolating from our results and expecting that models will behave ``pragmatically'' in downstream applications. This is especially true for models behind the OpenAI API, and \davinci{} in particular, for which very little is publicly known about the training protocol. 

\section*{Ethics statement} 

Language technologies have the potential to cause harm at the individual and societal levels. Large language models (LLMs), which are typically trained on vast amounts of internet text, have been shown to perpetuate stereotypes based on gender, race, and sexual orientation. Applications using LLMs could reinforce systematic discrimination and amplify existing socioeconomic inequities. For example, LLMs could perpetuate social biases by assisting with hiring decisions or legal rulings. 

The remarkable fluency of LLM-generated text also poses risks for the general public. LLMs have long been used to generate text that is difficult to distinguish from human-written text, raising concerns about detecting fake news and misinformation. Recently, LLMs have been used to synthesize knowledge -- for example, by answering scientific questions \citep{taylor_galactica_2022} or acting as search engines \citep{shah_situating_2022}. Using LLMs as knowledge-providers could tremendously impact the nature of human collaboration and work, raising the need for model transparency and explainability.

\bibliography{custom}

\begin{thebibliography}{100}
\expandafter\ifx\csname natexlab\endcsname\relax\def\natexlab#1{#1}\fi

\bibitem[{Ameka and Terkourafi(2019)}]{ameka_what_2019}
Felix~K. Ameka and Marina Terkourafi. 2019.
\newblock \href {https://doi.org/https://doi.org/10.1016/j.pragma.2019.04.001}
  {What if…? {Imagining} non-{Western} perspectives on pragmatic theory and
  practice}.
\newblock \emph{Journal of Pragmatics}, 145:72--82.

\bibitem[{Andreas(2022)}]{andreas_language_2022}
Jacob Andreas. 2022.
\newblock \href {https://aclanthology.org/2022.findings-emnlp.423} {Language
  {Models} as {Agent} {Models}}.
\newblock In \emph{Findings of the {Association} for {Computational}
  {Linguistics}: {EMNLP} 2022}, pages 5769--5779, Abu Dhabi, United Arab
  Emirates. Association for Computational Linguistics.

\bibitem[{Andrés-Roqueta and Katsos(2017)}]{andres-roqueta_contribution_2017}
Clara Andrés-Roqueta and Napoleon Katsos. 2017.
\newblock \href {https://doi.org/10.3389/fpsyg.2017.00996} {The {Contribution}
  of {Grammar}, {Vocabulary} and {Theory} of {Mind} in {Pragmatic} {Language}
  {Competence} in {Children} with {Autistic} {Spectrum} {Disorders}}.
\newblock \emph{Frontiers in Psychology}, 8.

\bibitem[{Apperly(2011)}]{apperly_mindreaders_2011}
Ian Apperly. 2011.
\newblock \emph{Mindreaders: {The} cognitive basis of "{Theory} of {Mind}"}.
\newblock Psychology Press, New York.

\bibitem[{Attardo(2000)}]{attardo_irony_2000}
Salvatore Attardo. 2000.
\newblock \href {https://doi.org/10.1016/S0378-2166(99)00070-3} {Irony as
  relevant inappropriateness}.
\newblock \emph{Journal of Pragmatics}, 32(6):793--826.

\bibitem[{Austin(1975)}]{austin_how_1975}
John~L. Austin. 1975.
\newblock \emph{How to do things with words}.

\bibitem[{Aykan and Nalçacı(2018)}]{aykan_assessing_2018}
Simge Aykan and Erhan Nalçacı. 2018.
\newblock \href {https://doi.org/10.3389/fpsyg.2018.01470} {Assessing {Theory}
  of {Mind} by {Humor}: {The} {Humor} {Comprehension} and {Appreciation} {Test}
  ({ToM}-{HCAT})}.
\newblock \emph{Frontiers in Psychology}, 9.

\bibitem[{Baron-Cohen et~al.(1985)Baron-Cohen, Leslie, and
  Frith}]{baron-cohen_does_1985}
Simon Baron-Cohen, Alan~M. Leslie, and Uta Frith. 1985.
\newblock \href {https://doi.org/10.1016/0010-0277(85)90022-8} {Does the
  autistic child have a “theory of mind”?}
\newblock \emph{Cognition}, 21(1):37--46.

\bibitem[{Bender and Koller(2020)}]{bender_climbing_2020}
Emily~M. Bender and Alexander Koller. 2020.
\newblock \href {https://doi.org/10.18653/v1/2020.acl-main.463} {Climbing
  towards {NLU}: {On} {Meaning}, {Form}, and {Understanding} in the {Age} of
  {Data}}.
\newblock In \emph{Proceedings of the 58th {Annual} {Meeting} of the
  {Association} for {Computational} {Linguistics}}, pages 5185--5198, Online.
  Association for Computational Linguistics.

\bibitem[{Bergen et~al.(2016)Bergen, Levy, and Goodman}]{bergen_pragmatic_2016}
Leon Bergen, Roger Levy, and Noah~D. Goodman. 2016.
\newblock Pragmatic reasoning through semantic inference.
\newblock \emph{Semantics and Pragmatics}, 9.

\bibitem[{Beyer et~al.(2021)Beyer, Loáiciga, and Schlangen}]{beyer_is_2021}
Anne Beyer, Sharid Loáiciga, and David Schlangen. 2021.
\newblock \href {https://doi.org/10.18653/v1/2021.naacl-main.328} {Is
  {Incoherence} {Surprising}? {Targeted} {Evaluation} of {Coherence}
  {Prediction} from {Language} {Models}}.
\newblock In \emph{Proceedings of the 2021 {Conference} of the {North}
  {American} {Chapter} of the {Association} for {Computational} {Linguistics}:
  {Human} {Language} {Technologies}}, pages 4164--4173, Online. Association for
  Computational Linguistics.

\bibitem[{Bischetti et~al.(2019)Bischetti, Ceccato, Lecce, Cavallini, and
  Bambini}]{bischetti_pragmatics_2019}
Luca Bischetti, Irene Ceccato, Serena Lecce, Elena Cavallini, and Valentina
  Bambini. 2019.
\newblock \href {https://doi.org/10.1007/s12144-019-00295-w} {Pragmatics and
  theory of mind in older adults’ humor comprehension}.
\newblock \emph{Current Psychology}.

\bibitem[{Bisk et~al.(2020)Bisk, Holtzman, Thomason, Andreas, Bengio, Chai,
  Lapata, Lazaridou, May, Nisnevich, Pinto, and Turian}]{bisk_experience_2020}
Yonatan Bisk, Ari Holtzman, Jesse Thomason, Jacob Andreas, Yoshua Bengio, Joyce
  Chai, Mirella Lapata, Angeliki Lazaridou, Jonathan May, Aleksandr Nisnevich,
  Nicolas Pinto, and Joseph Turian. 2020.
\newblock \href {https://doi.org/10.18653/v1/2020.emnlp-main.703} {Experience
  {Grounds} {Language}}.
\newblock In \emph{Proceedings of the 2020 {Conference} on {Empirical}
  {Methods} in {Natural} {Language} {Processing} ({EMNLP})}, pages 8718--8735,
  Online. Association for Computational Linguistics.

\bibitem[{Blasi et~al.(2022)Blasi, Henrich, Adamou, Kemmerer, and
  Majid}]{blasi_over-reliance_2022}
Damián~E. Blasi, Joseph Henrich, Evangelia Adamou, David Kemmerer, and Asifa
  Majid. 2022.
\newblock \href {https://doi.org/10.1016/j.tics.2022.09.015} {Over-reliance on
  {English} hinders cognitive science}.
\newblock \emph{Trends in Cognitive Sciences}, 26(12):1153--1170.
\newblock Publisher: Elsevier.

\bibitem[{Booth(1974)}]{booth_rhetoric_1974}
W.C. Booth. 1974.
\newblock \href {https://books.google.com/books?id=jbgufPEUD6QC} {\emph{A
  {Rhetoric} of {Irony}}}.
\newblock Literature/{Criticism} - {The} {University} of {Chicago} {Press}.
  University of Chicago Press.

\bibitem[{Brennan et~al.(2010)Brennan, Galati, and Kuhlen}]{brennan_two_2010}
Susan~E. Brennan, Alexia Galati, and Anna~K. Kuhlen. 2010.
\newblock \href {https://doi.org/10.1016/S0079-7421(10)53008-1} {Two {Minds},
  {One} {Dialog}: {Coordinating} {Speaking} and {Understanding}}.
\newblock In Brian~H. Ross, editor, \emph{Psychology of {Learning} and
  {Motivation}}, volume~53, pages 301--344. Academic Press.

\bibitem[{Brown and Levinson(1987)}]{brown_politeness_1987}
Penelope Brown and Stephen~C. Levinson. 1987.
\newblock \emph{Politeness: {Some} {Universals} in {Language} {Usage}}.
\newblock Cambridge University Press.

\bibitem[{Brown et~al.(2021)Brown, Sicoli, and Guen}]{brown_cross-speaker_2021}
Penelope Brown, Mark~A. Sicoli, and Olivier~Le Guen. 2021.
\newblock \href {https://doi.org/https://doi.org/10.1016/j.pragma.2021.07.005}
  {Cross-speaker repetition and epistemic stance in {Tzeltal}, {Yucatec}, and
  {Zapotec} conversations}.
\newblock \emph{Journal of Pragmatics}, 183:256--272.

\bibitem[{Butterfill and Apperly(2013)}]{butterfill_how_2013}
Stephen~A. Butterfill and Ian~A. Apperly. 2013.
\newblock \href {https://doi.org/10.1111/mila.12036} {How to {Construct} a
  {Minimal} {Theory} of {Mind}}.
\newblock \emph{Mind \& Language}, 28(5):606--637.
\newblock Publisher: John Wiley \& Sons, Ltd.

\bibitem[{Camden et~al.(1984)Camden, Motley, and Wilson}]{camden_white_1984}
Carl Camden, Michael~T. Motley, and Ann Wilson. 1984.
\newblock \href {https://doi.org/10.1080/10570318409374167} {White lies in
  interpersonal communication: {A} taxonomy and preliminary investigation of
  social motivations}.
\newblock \emph{Western Journal of Speech Communication}, 48(4):309--325.
\newblock Publisher: Routledge.

\bibitem[{Card et~al.(2020)Card, Henderson, Khandelwal, Jia, Mahowald, and
  Jurafsky}]{card_little_2020}
Dallas Card, Peter Henderson, Urvashi Khandelwal, Robin Jia, Kyle Mahowald, and
  Dan Jurafsky. 2020.
\newblock \href {https://doi.org/10.18653/v1/2020.emnlp-main.745} {With
  {Little} {Power} {Comes} {Great} {Responsibility}}.
\newblock In \emph{Proceedings of the 2020 {Conference} on {Empirical}
  {Methods} in {Natural} {Language} {Processing} ({EMNLP})}, pages 9263--9274,
  Online. Association for Computational Linguistics.

\bibitem[{Chowdhery et~al.(2022)Chowdhery, Narang, Devlin, Bosma, Mishra,
  Roberts, Barham, Chung, Sutton, Gehrmann, Schuh, Shi, Tsvyashchenko, Maynez,
  Rao, Barnes, Tay, Shazeer, Prabhakaran, Reif, Du, Hutchinson, Pope, Bradbury,
  Austin, Isard, Gur-Ari, Yin, Duke, Levskaya, Ghemawat, Dev, Michalewski,
  Garcia, Misra, Robinson, Fedus, Zhou, Ippolito, Luan, Lim, Zoph, Spiridonov,
  Sepassi, Dohan, Agrawal, Omernick, Dai, Pillai, Pellat, Lewkowycz, Moreira,
  Child, Polozov, Lee, Zhou, Wang, Saeta, Diaz, Firat, Catasta, Wei,
  Meier-Hellstern, Eck, Dean, Petrov, and Fiedel}]{chowdhery_palm_2022}
Aakanksha Chowdhery, Sharan Narang, Jacob Devlin, Maarten Bosma, Gaurav Mishra,
  Adam Roberts, Paul Barham, Hyung~Won Chung, Charles Sutton, Sebastian
  Gehrmann, Parker Schuh, Kensen Shi, Sasha Tsvyashchenko, Joshua Maynez,
  Abhishek Rao, Parker Barnes, Yi~Tay, Noam Shazeer, Vinodkumar Prabhakaran,
  Emily Reif, Nan Du, Ben Hutchinson, Reiner Pope, James Bradbury, Jacob
  Austin, Michael Isard, Guy Gur-Ari, Pengcheng Yin, Toju Duke, Anselm
  Levskaya, Sanjay Ghemawat, Sunipa Dev, Henryk Michalewski, Xavier Garcia,
  Vedant Misra, Kevin Robinson, Liam Fedus, Denny Zhou, Daphne Ippolito, David
  Luan, Hyeontaek Lim, Barret Zoph, Alexander Spiridonov, Ryan Sepassi, David
  Dohan, Shivani Agrawal, Mark Omernick, Andrew~M. Dai,
  Thanumalayan~Sankaranarayana Pillai, Marie Pellat, Aitor Lewkowycz, Erica
  Moreira, Rewon Child, Oleksandr Polozov, Katherine Lee, Zongwei Zhou, Xuezhi
  Wang, Brennan Saeta, Mark Diaz, Orhan Firat, Michele Catasta, Jason Wei,
  Kathy Meier-Hellstern, Douglas Eck, Jeff Dean, Slav Petrov, and Noah Fiedel.
  2022.
\newblock \href {https://arxiv.org/abs/2204.02311} {{PaLM}: {Scaling}
  {Language} {Modeling} with {Pathways}}.
\newblock {arXiv} preprint.

\bibitem[{Chung et~al.(2022)Chung, Hou, Longpre, Zoph, Tay, Fedus, Li, Wang,
  Dehghani, Brahma, Webson, Gu, Dai, Suzgun, Chen, Chowdhery, Narang, Mishra,
  Yu, Zhao, Huang, Dai, Yu, Petrov, Chi, Dean, Devlin, Roberts, Zhou, Le, and
  Wei}]{chung_scaling_2022}
Hyung~Won Chung, Le~Hou, Shayne Longpre, Barret Zoph, Yi~Tay, William Fedus,
  Eric Li, Xuezhi Wang, Mostafa Dehghani, Siddhartha Brahma, Albert Webson,
  Shixiang~Shane Gu, Zhuyun Dai, Mirac Suzgun, Xinyun Chen, Aakanksha
  Chowdhery, Sharan Narang, Gaurav Mishra, Adams Yu, Vincent Zhao, Yanping
  Huang, Andrew Dai, Hongkun Yu, Slav Petrov, Ed~H. Chi, Jeff Dean, Jacob
  Devlin, Adam Roberts, Denny Zhou, Quoc~V. Le, and Jason Wei. 2022.
\newblock \href {https://arxiv.org/abs/2210.11416} {Scaling
  {Instruction}-{Finetuned} {Language} {Models}}.
\newblock {arXiv} preprint.

\bibitem[{Dasgupta et~al.(2018)Dasgupta, Guo, Stuhlmüller, Gershman, and
  Goodman}]{dasgupta_evaluating_2018}
Ishita Dasgupta, Demi Guo, Andreas Stuhlmüller, Samuel~J. Gershman, and
  Noah~D. Goodman. 2018.
\newblock \href {https://arxiv.org/abs/1802.04302} {Evaluating
  {Compositionality} in {Sentence} {Embeddings}}.
\newblock In \emph{Proceedings of the {Cognitive} {Science} {Society}}.

\bibitem[{Deckers and Kizer(1975)}]{deckers_humor_1975}
Lambert Deckers and Philip Kizer. 1975.
\newblock \href {https://doi.org/10.1080/00223980.1975.9915778} {Humor and the
  {Incongruity} {Hypothesis}}.
\newblock \emph{The Journal of Psychology}, 90(2):215--218.

\bibitem[{Degen(2023)}]{degen_rational_2023}
Judith Degen. 2023.
\newblock \href {https://doi.org/10.1146/annurev-linguistics-031220-010811}
  {The {Rational} {Speech} {Act} {Framework}}.
\newblock \emph{Annual Review of Linguistics}, 9(1):519--540.

\bibitem[{Demedardi et~al.(2021)Demedardi, Brechet, Gentaz, and
  Monnier}]{demedardi_prosocial_2021}
Marie-Julie Demedardi, Claire Brechet, Edouard Gentaz, and Catherine Monnier.
  2021.
\newblock \href {https://doi.org/10.1016/j.jecp.2020.105045} {Prosocial lying
  in children between 4 and 11 years of age: {The} role of emotional
  understanding and empathy}.
\newblock \emph{Journal of Experimental Child Psychology}, 203:105045.

\bibitem[{Dideriksen et~al.(2022)Dideriksen, Christiansen, Dingemanse,
  Højmark-Bertelsen, Johansson, Tylén, and
  Fusaroli}]{dideriksen_language_2022}
Christina Dideriksen, Morten~H Christiansen, Mark Dingemanse, Malte
  Højmark-Bertelsen, Christer Johansson, Kristian Tylén, and Riccardo
  Fusaroli. 2022.
\newblock \href {https://doi.org/10.31234/osf.io/t3s6c} {Language specific
  constraints on conversation: {Evidence} from {Danish} and {Norwegian}}.
\newblock PsyArXiv preprint.

\bibitem[{Dombi et~al.(2022)Dombi, Sydorenko, and
  Timpe-Laughlin}]{dombi_common_2022}
Judit Dombi, Tetyana Sydorenko, and Veronika Timpe-Laughlin. 2022.
\newblock \href {https://doi.org/10.1016/j.pragma.2022.03.001} {Common ground,
  cooperation, and recipient design in human-computer interactions}.
\newblock \emph{Journal of Pragmatics}, 193:4--20.

\bibitem[{Enrici et~al.(2019)Enrici, Bara, and Adenzato}]{enrici_theory_2019}
Ivan Enrici, Bruno~G. Bara, and Mauro Adenzato. 2019.
\newblock \href {https://doi.org/https://doi.org/10.1075/pc.19010.enr} {Theory
  of {Mind}, pragmatics and the brain: {Converging} evidence for the role of
  intention processing as a core feature of human communication}.
\newblock \emph{Pragmatics \& Cognition}, 26(1):5--38.

\bibitem[{Floyd et~al.(In prep)Floyd, Jouravlev, Mineroff, Bergen, Fedorenko,
  and Gibson}]{floyd_deciphering_nodate}
Sammy Floyd, Olessia Jouravlev, Zachary Mineroff, Leon Bergen, Evelina
  Fedorenko, and Edward Gibson. In prep.
\newblock Deciphering the structure of pragmatics: {A} large-scale individual
  differences investigation.

\bibitem[{Floyd(2021)}]{floyd_conversation_2021}
Simeon Floyd. 2021.
\newblock \href {https://doi.org/10.1146/annurev-anthro-101819-110158}
  {Conversation and {Culture}}.
\newblock \emph{Annual Review of Anthropology}, 50(1):219--240.
\newblock Publisher: Annual Reviews.

\bibitem[{Frank et~al.(2018)Frank, Emilsson, Peloquin, Goodman, and
  Potts}]{frank_rational_2018}
Michael~C Frank, Andrés~Goméz Emilsson, Benjamin Peloquin, Noah~D. Goodman,
  and Christopher Potts. 2018.
\newblock \href {https://psyarxiv.com/f9y6b} {Rational speech act models of
  pragmatic reasoning in reference games}.
\newblock PsyArXiv preprint.

\bibitem[{Frank and Goodman(2012)}]{frank_predicting_2012}
Michael~C. Frank and Noah~D. Goodman. 2012.
\newblock \href {https://doi.org/10.1126/science.1218633} {Predicting
  {Pragmatic} {Reasoning} in {Language} {Games}}.
\newblock \emph{Science}, 336(6084):998--998.

\bibitem[{Franke and Jäger(2016)}]{franke_probabilistic_2016}
Michael Franke and Gerhard Jäger. 2016.
\newblock \href {https://doi.org/doi:10.1515/zfs-2016-0002} {Probabilistic
  pragmatics, or why {Bayes}’ rule is probably important for pragmatics}.
\newblock \emph{Zeitschrift für Sprachwissenschaft}, 35(1):3--44.

\bibitem[{Futrell et~al.(2019)Futrell, Wilcox, Morita, Qian, Ballesteros, and
  Levy}]{futrell_neural_2019}
Richard Futrell, Ethan Wilcox, Takashi Morita, Peng Qian, Miguel Ballesteros,
  and Roger Levy. 2019.
\newblock \href {https://doi.org/10.18653/v1/N19-1004} {Neural language models
  as psycholinguistic subjects: {Representations} of syntactic state}.
\newblock In \emph{Proceedings of the 2019 {Conference} of the {North}
  {American} {Chapter} of the {Association} for {Computational} {Linguistics}:
  {Human} {Language} {Technologies}, {Volume} 1 ({Long} and {Short} {Papers})},
  pages 32--42, Minneapolis, Minnesota. Association for Computational
  Linguistics.

\bibitem[{Gibbs(1979)}]{gibbs_contextual_1979}
Raymond~W. Gibbs. 1979.
\newblock \href {https://doi.org/10.1080/01638537909544450} {Contextual effects
  in understanding indirect requests}.
\newblock \emph{Discourse Processes}, 2(1):1--10.
\newblock Publisher: Routledge.

\bibitem[{Grice(1975)}]{grice_logic_1975}
Herbert~P. Grice. 1975.
\newblock \href {http://www.ucl.ac.uk/ls/studypacks/Grice-Logic.pdf} {Logic and
  {Conversation}}.
\newblock In Peter Cole and Jerry~L. Morgan, editors, \emph{Syntax and
  {Semantics}: {Speech} {Acts}}, volume~3, pages 41--58. Academic Press.

\bibitem[{Gu et~al.(2022)Gu, Fu, Pyatkin, Magnusson, Dalvi~Mishra, and
  Clark}]{gu_just-dream-about-it_2022}
Yuling Gu, Yao Fu, Valentina Pyatkin, Ian Magnusson, Bhavana Dalvi~Mishra, and
  Peter Clark. 2022.
\newblock \href {https://aclanthology.org/2022.flp-1.12}
  {Just-{DREAM}-about-it: {Figurative} {Language} {Understanding} with
  {DREAM}-{FLUTE}}.
\newblock In \emph{Proceedings of the 3rd {Workshop} on {Figurative} {Language}
  {Processing} ({FLP})}, pages 84--93, Abu Dhabi, United Arab Emirates
  (Hybrid). Association for Computational Linguistics.

\bibitem[{Happé(1993)}]{happe_communicative_1993}
Francesca~G.E. Happé. 1993.
\newblock \href {https://doi.org/10.1016/0010-0277(93)90026-R} {Communicative
  competence and theory of mind in autism: {A} test of relevance theory}.
\newblock \emph{Cognition}, 48(2):101--119.

\bibitem[{Heyes(2014)}]{heyes_submentalizing_2014}
Cecilia Heyes. 2014.
\newblock \href {https://doi.org/10.1177/1745691613518076} {Submentalizing: {I}
  {Am} {Not} {Really} {Reading} {Your} {Mind}}.
\newblock \emph{Perspectives on Psychological Science}, 9(2):131--143.

\bibitem[{Horn(1972)}]{horn_semantic_1972}
Laurence~R. Horn. 1972.
\newblock \href {https://linguistics.ucla.edu/images/stories/Horn.1972.pdf}
  {\emph{On the semantic properties of logical operators in {English}}}.
\newblock {PhD} {Thesis}, University of California Los Angeles.

\bibitem[{Hsu and Cheung(2013)}]{hsu_two_2013}
Yik~Kwan Hsu and Him Cheung. 2013.
\newblock \href {https://doi.org/10.1037/a0031128} {Two mentalizing capacities
  and the understanding of two types of lie telling in children}.
\newblock \emph{Developmental Psychology}, 49:1650--1659.

\bibitem[{Hu et~al.(2020)Hu, Gauthier, Qian, Wilcox, and
  Levy}]{hu_systematic_2020}
Jennifer Hu, Jon Gauthier, Peng Qian, Ethan Wilcox, and Roger Levy. 2020.
\newblock \href {https://doi.org/10.18653/v1/2020.acl-main.158} {A {Systematic}
  {Assessment} of {Syntactic} {Generalization} in {Neural} {Language}
  {Models}}.
\newblock In \emph{Proceedings of the 58th {Annual} {Meeting} of the
  {Association} for {Computational} {Linguistics}}, pages 1725--1744, Online.
  Association for Computational Linguistics.

\bibitem[{Jacoby and Fedorenko(2020)}]{jacoby_discourse-level_2020}
Nir Jacoby and Evelina Fedorenko. 2020.
\newblock \href {https://doi.org/10.1080/23273798.2018.1525494}
  {Discourse-level comprehension engages medial frontal {Theory} of {Mind}
  brain regions even for expository texts}.
\newblock \emph{Language, Cognition and Neuroscience}, 35(6):780--796.

\bibitem[{Jeretic et~al.(2020)Jeretic, Warstadt, Bhooshan, and
  Williams}]{jeretic_are_2020}
Paloma Jeretic, Alex Warstadt, Suvrat Bhooshan, and Adina Williams. 2020.
\newblock \href {https://doi.org/10.18653/v1/2020.acl-main.768} {Are {Natural}
  {Language} {Inference} {Models} {IMPPRESsive}? {Learning} {IMPlicature} and
  {PRESupposition}}.
\newblock In \emph{Proceedings of the 58th {Annual} {Meeting} of the
  {Association} for {Computational} {Linguistics}}, pages 8690--8705, Online.
  Association for Computational Linguistics.

\bibitem[{Kao et~al.(2014)Kao, Bergen, and Goodman}]{kao_formalizing_2014}
Justine~T. Kao, Leon Bergen, and Noah~D. Goodman. 2014.
\newblock \href {https://escholarship.org/uc/item/09h3p4cz} {Formalizing the
  {Pragmatics} of {Metaphor} {Understanding}}.
\newblock In \emph{Proceedings of the 36th {Annual} {Meeting} of the
  {Cognitive} {Science} {Society}}.

\bibitem[{Kao and Goodman(2014)}]{kao_lets_2014}
Justine~T. Kao and Noah~D. Goodman. 2014.
\newblock \href {https://cocolab.stanford.edu/papers/KaoEtAl2015-Cogsci.pdf}
  {Let's talk (ironically) about the weather: {Modeling} verbal irony}.
\newblock In \emph{Proceedings of the 36th {Annual} {Meeting} of the
  {Cognitive} {Science} {Society}}.

\bibitem[{Kline~Struhl et~al.(2018)Kline~Struhl, Gallée, Balewski, and
  Fedorenko}]{kline_struhl_understanding_2018}
Melissa Kline~Struhl, Jeanne Gallée, Zuzanna Balewski, and Evelina Fedorenko.
  2018.
\newblock \href {https://psyarxiv.com/h2nyx} {Understanding jokes draws most
  heavily on the {Theory} of {Mind} brain network}.
\newblock PsyArXiv preprint.

\bibitem[{Kosinski(2023)}]{kosinski_theory_2023}
Michal Kosinski. 2023.
\newblock \href {https://arxiv.org/abs/2302.02083} {Theory of {Mind} {May}
  {Have} {Spontaneously} {Emerged} in {Large} {Language} {Models}}.
\newblock {arXiv} preprint.

\bibitem[{Kreiss et~al.(2022)Kreiss, Fang, Goodman, and
  Potts}]{kreiss-etal-2022-concadia}
Elisa Kreiss, Fei Fang, Noah Goodman, and Christopher Potts. 2022.
\newblock \href {https://aclanthology.org/2022.emnlp-main.308} {Concadia:
  Towards image-based text generation with a purpose}.
\newblock In \emph{Proceedings of the 2022 Conference on Empirical Methods in
  Natural Language Processing}, pages 4667--4684, Abu Dhabi, United Arab
  Emirates. Association for Computational Linguistics.

\bibitem[{Lakoff and Johnson(1980)}]{lakoff_metaphors_1980}
G.~Lakoff and M.~Johnson. 1980.
\newblock \href
  {https://press.uchicago.edu/ucp/books/book/chicago/M/bo3637992.html}
  {\emph{Metaphors {We} {Live} {By}}}.
\newblock University of Chicago Press.

\bibitem[{Le et~al.(2019)Le, Boureau, and Nickel}]{le_revisiting_2019}
Matthew Le, Y-Lan Boureau, and Maximilian Nickel. 2019.
\newblock \href {https://doi.org/10.18653/v1/D19-1598} {Revisiting the
  {Evaluation} of {Theory} of {Mind} through {Question} {Answering}}.
\newblock In \emph{Proceedings of the 2019 {Conference} on {Empirical}
  {Methods} in {Natural} {Language} {Processing} and the 9th {International}
  {Joint} {Conference} on {Natural} {Language} {Processing}
  ({EMNLP}-{IJCNLP})}, pages 5872--5877, Hong Kong, China. Association for
  Computational Linguistics.

\bibitem[{Leslie et~al.(2004)Leslie, Friedman, and German}]{leslie_core_2004}
Alan~M. Leslie, Ori Friedman, and Tim~P. German. 2004.
\newblock \href {https://doi.org/10.1016/j.tics.2004.10.001} {Core mechanisms
  in ‘theory of mind’}.
\newblock \emph{Trends in Cognitive Sciences}, 8(12):528--533.

\bibitem[{Levinson(2000)}]{levinson_presumptive_2000}
Stephen Levinson. 2000.
\newblock \emph{Presumptive meaning: {The} theory of generalized conversational
  implicature}.
\newblock MIT Press.

\bibitem[{Li et~al.(2021)Li, Schuster, and Degen}]{li_predicting_2021}
Elissa Li, Sebastian Schuster, and Judith Degen. 2021.
\newblock \href {https://doi.org/10.7275/xr01-a852} {Predicting {Scalar}
  {Inferences} {From} "{Or}" to "{Not} {Both}" {Using} {Neural} {Sentence}
  {Encoders}}.
\newblock In \emph{Proceedings of the {Society} for {Computation} in
  {Linguistics}}, volume~4.

\bibitem[{Li(2012)}]{li_cultural_2012}
Jin Li. 2012.
\newblock \href {https://doi.org/10.1017/CBO9781139028400} {\emph{Cultural
  {Foundations} of {Learning}: {East} and {West}}}.
\newblock Cambridge University Press, Cambridge.

\bibitem[{Linzen et~al.(2016)Linzen, Dupoux, and
  Goldberg}]{linzen_assessing_2016}
Tal Linzen, Emmanuel Dupoux, and Yoav Goldberg. 2016.
\newblock \href {https://doi.org/10.1162/tacl_a_00115} {Assessing the {Ability}
  of {LSTMs} to {Learn} {Syntax}-{Sensitive} {Dependencies}}.
\newblock \emph{Transactions of the Association for Computational Linguistics},
  4:521--535.
\newblock Place: Cambridge, MA Publisher: MIT Press.

\bibitem[{Liu et~al.(2022)Liu, Cui, Zheng, and Neubig}]{liu_testing_2022}
Emmy Liu, Chenxuan Cui, Kenneth Zheng, and Graham Neubig. 2022.
\newblock \href {https://doi.org/10.18653/v1/2022.naacl-main.330} {Testing the
  {Ability} of {Language} {Models} to {Interpret} {Figurative} {Language}}.
\newblock In \emph{Proceedings of the 2022 {Conference} of the {North}
  {American} {Chapter} of the {Association} for {Computational} {Linguistics}:
  {Human} {Language} {Technologies}}, pages 4437--4452, Seattle, United States.
  Association for Computational Linguistics.

\bibitem[{Lumer and Buschmeier(2022)}]{lumer_modeling_2022}
Eleonore Lumer and Hendrik Buschmeier. 2022.
\newblock \href {https://escholarship.org/uc/item/7qg325fr} {Modeling {Social}
  {Influences} on {Indirectness} in a {Rational} {Speech} {Act} {Approach} to
  {Politeness}}.
\newblock In \emph{Proceedings of the 44th {Annual} {Conference} of the
  {Cognitive} {Science} {Society}}.

\bibitem[{Mahowald et~al.(2023)Mahowald, Ivanova, Blank, Kanwisher, Tenenbaum,
  and Fedorenko}]{mahowald_dissociating_2023}
Kyle Mahowald, Anna~A. Ivanova, Idan~A. Blank, Nancy Kanwisher, Joshua~B.
  Tenenbaum, and Evelina Fedorenko. 2023.
\newblock \href {https://arxiv.org/abs/2301.06627} {Dissociating language and
  thought in large language models: {A} cognitive perspective}.
\newblock {arXiv} preprint.

\bibitem[{Martin and Ford(2018)}]{martin_psychology_2018}
R.A. Martin and T.~Ford. 2018.
\newblock \emph{The {Psychology} of {Humor}: {An} {Integrative} {Approach}}.
\newblock Academic Press.

\bibitem[{McCoy et~al.(2019)McCoy, Pavlick, and Linzen}]{mccoy_right_2019}
Tom McCoy, Ellie Pavlick, and Tal Linzen. 2019.
\newblock \href {https://doi.org/10.18653/v1/P19-1334} {Right for the {Wrong}
  {Reasons}: {Diagnosing} {Syntactic} {Heuristics} in {Natural} {Language}
  {Inference}}.
\newblock In \emph{Proceedings of the 57th {Annual} {Meeting} of the
  {Association} for {Computational} {Linguistics}}, pages 3428--3448, Florence,
  Italy. Association for Computational Linguistics.

\bibitem[{Michael(2020)}]{michael_dissect_2020}
Julian Michael. 2020.
\newblock \href
  {https://julianmichael.org/blog/2020/07/23/to-dissect-an-octopus.html} {To
  {Dissect} an {Octopus}: {Making} {Sense} of the {Form}/{Meaning} {Debate}}.

\bibitem[{Moss and Schunn(2015)}]{moss_comprehension_2015}
Jarrod Moss and Christian~D. Schunn. 2015.
\newblock \href {https://doi.org/10.3389/fnhum.2015.00562} {Comprehension
  through explanation as the interaction of the brain’s coherence and
  cognitive control networks}.
\newblock \emph{Frontiers in Human Neuroscience}, 9.

\bibitem[{Nematzadeh et~al.(2018)Nematzadeh, Burns, Grant, Gopnik, and
  Griffiths}]{nematzadeh_evaluating_2018}
Aida Nematzadeh, Kaylee Burns, Erin Grant, Alison Gopnik, and Tom Griffiths.
  2018.
\newblock \href {https://doi.org/10.18653/v1/D18-1261} {Evaluating {Theory} of
  {Mind} in {Question} {Answering}}.
\newblock In \emph{Proceedings of the 2018 {Conference} on {Empirical}
  {Methods} in {Natural} {Language} {Processing}}, pages 2392--2400, Brussels,
  Belgium. Association for Computational Linguistics.

\bibitem[{Nie et~al.(2019)Nie, Wang, and Bansal}]{nie_analyzing_2019}
Yixin Nie, Yicheng Wang, and Mohit Bansal. 2019.
\newblock \href {https://doi.org/10.1609/aaai.v33i01.33016867} {Analyzing
  {Compositionality}-{Sensitivity} of {NLI} {Models}}.
\newblock \emph{Proceedings of the AAAI Conference on Artificial Intelligence},
  33(01):6867--6874.

\bibitem[{Ouyang et~al.(2022)Ouyang, Wu, Jiang, Almeida, Wainwright, Mishkin,
  Zhang, Agarwal, Slama, Ray, Schulman, Hilton, Kelton, Miller, Simens, Askell,
  Welinder, Christiano, Leike, and Lowe}]{ouyang_training_2022}
Long Ouyang, Jeff Wu, Xu~Jiang, Diogo Almeida, Carroll~L. Wainwright, Pamela
  Mishkin, Chong Zhang, Sandhini Agarwal, Katarina Slama, Alex Ray, John
  Schulman, Jacob Hilton, Fraser Kelton, Luke Miller, Maddie Simens, Amanda
  Askell, Peter Welinder, Paul Christiano, Jan Leike, and Ryan Lowe. 2022.
\newblock \href {https://doi.org/10.48550/ARXIV.2203.02155} {Training language
  models to follow instructions with human feedback}.

\bibitem[{Potts(2020)}]{potts_is_2020}
Christopher Potts. 2020.
\newblock \href
  {https://chrisgpotts.medium.com/is-it-possible-for-language-models-to-achieve-language-understanding-81df45082ee2}
  {Is it possible for language models to achieve language understanding?}

\bibitem[{Potts et~al.(2016)Potts, Lassiter, Levy, and
  Frank}]{potts_embedded_2016}
Christopher Potts, Daniel Lassiter, Roger Levy, and Michael~C. Frank. 2016.
\newblock \href {https://doi.org/10.1093/jos/ffv012} {Embedded {Implicatures}
  as {Pragmatic} {Inferences} under {Compositional} {Lexical} {Uncertainty}}.
\newblock \emph{Journal of Semantics}, 33(4):755--802.

\bibitem[{Radford et~al.(2019)Radford, Wu, Child, Luan, Amodei, and
  Sutskever}]{radford_language_2019}
Alec Radford, Jeff Wu, Rewon Child, David Luan, Dario Amodei, and Ilya
  Sutskever. 2019.
\newblock \href
  {https://d4mucfpksywv.cloudfront.net/better-language-models/language-models.pdf}
  {Language {Models} are {Unsupervised} {Multitask} {Learners}}.

\bibitem[{Raffel et~al.(2020)Raffel, Shazeer, Roberts, Lee, Narang, Matena,
  Zhou, Li, and Liu}]{raffel_exploring_2020}
Colin Raffel, Noam Shazeer, Adam Roberts, Katherine Lee, Sharan Narang, Michael
  Matena, Yanqi Zhou, Wei Li, and Peter~J. Liu. 2020.
\newblock \href {http://jmlr.org/papers/v21/20-074.html} {Exploring the
  {Limits} of {Transfer} {Learning} with a {Unified} {Text}-to-{Text}
  {Transformer}}.
\newblock \emph{Journal of Machine Learning Research}, 21(140):1--67.

\bibitem[{Rubio-Fernandez(2021)}]{rubio-fernandez_pragmatic_2021}
Paula Rubio-Fernandez. 2021.
\newblock \href {https://doi.org/10.1007/s11229-020-02768-z} {Pragmatic
  markers: the missing link between language and {Theory} of {Mind}}.
\newblock \emph{Synthese}, 199(1):1125--1158.

\bibitem[{Rubio-Fernandez and
  Jara-Ettinger(2020)}]{rubio-fernandez_incrementality_2020}
Paula Rubio-Fernandez and Julian Jara-Ettinger. 2020.
\newblock \href {https://doi.org/10.1073/pnas.1922067117} {Incrementality and
  efficiency shape pragmatics across languages}.
\newblock \emph{Proceedings of the National Academy of Sciences},
  117(24):13399--13404.

\bibitem[{Ruis et~al.(2022)Ruis, Khan, Biderman, Hooker, Rocktäschel, and
  Grefenstette}]{ruis_large_2022}
Laura Ruis, Akbir Khan, Stella Biderman, Sara Hooker, Tim Rocktäschel, and
  Edward Grefenstette. 2022.
\newblock \href {https://arxiv.org/abs/2210.14986} {Large language models are
  not zero-shot communicators}.
\newblock {arXiv} preprint.

\bibitem[{Sap et~al.(2022)Sap, Le~Bras, Fried, and Choi}]{sap_neural_2022}
Maarten Sap, Ronan Le~Bras, Daniel Fried, and Yejin Choi. 2022.
\newblock \href {https://aclanthology.org/2022.emnlp-main.248} {Neural
  {Theory}-of-{Mind}? {On} the {Limits} of {Social} {Intelligence} in {Large}
  {LMs}}.
\newblock In \emph{Proceedings of the 2022 {Conference} on {Empirical}
  {Methods} in {Natural} {Language} {Processing}}, pages 3762--3780, Abu Dhabi,
  United Arab Emirates. Association for Computational Linguistics.

\bibitem[{Sap et~al.(2019)Sap, Rashkin, Chen, Le~Bras, and
  Choi}]{sap_social_2019}
Maarten Sap, Hannah Rashkin, Derek Chen, Ronan Le~Bras, and Yejin Choi. 2019.
\newblock \href {https://doi.org/10.18653/v1/D19-1454} {Social {IQa}:
  {Commonsense} {Reasoning} about {Social} {Interactions}}.
\newblock In \emph{Proceedings of the 2019 {Conference} on {Empirical}
  {Methods} in {Natural} {Language} {Processing} and the 9th {International}
  {Joint} {Conference} on {Natural} {Language} {Processing}
  ({EMNLP}-{IJCNLP})}, pages 4463--4473, Hong Kong, China. Association for
  Computational Linguistics.

\bibitem[{Saygin and Cicekli(2002)}]{saygin_pragmatics_2002}
Ayse~Pinar Saygin and Ilyas Cicekli. 2002.
\newblock \href {https://doi.org/10.1016/S0378-2166(02)80001-7} {Pragmatics in
  human-computer conversations}.
\newblock \emph{Journal of Pragmatics}, 34(3):227--258.

\bibitem[{Schuster et~al.(2020)Schuster, Chen, and
  Degen}]{schuster_harnessing_2020}
Sebastian Schuster, Yuxing Chen, and Judith Degen. 2020.
\newblock \href {https://doi.org/10.18653/v1/2020.acl-main.479} {Harnessing the
  linguistic signal to predict scalar inferences}.
\newblock In \emph{Proceedings of the 58th {Annual} {Meeting} of the
  {Association} for {Computational} {Linguistics}}, pages 5387--5403, Online.
  Association for Computational Linguistics.

\bibitem[{Searle(1975)}]{searle_indirect_1975}
John~R. Searle. 1975.
\newblock \href {https://doi.org/10.1163/9789004368811_004} {Indirect {Speech}
  {Acts}}.
\newblock In \emph{Speech {Acts}}, pages 59--82. Brill, Leiden, The
  Netherlands.

\bibitem[{Shah and Bender(2022)}]{shah_situating_2022}
Chirag Shah and Emily~M. Bender. 2022.
\newblock \href {https://doi.org/10.1145/3498366.3505816} {Situating {Search}}.
\newblock In \emph{{ACM} {SIGIR} {Conference} on {Human} {Information}
  {Interaction} and {Retrieval}}, {CHIIR} '22, pages 221--232, New York, NY,
  USA. Association for Computing Machinery.
\newblock Event-place: Regensburg, Germany.

\bibitem[{Spotorno et~al.(2012)Spotorno, Koun, Prado, Van Der~Henst, and
  Noveck}]{spotorno_neural_2012}
Nicola Spotorno, Eric Koun, Jérôme Prado, Jean-Baptiste Van Der~Henst, and
  Ira~A. Noveck. 2012.
\newblock \href {https://doi.org/10.1016/j.neuroimage.2012.06.046} {Neural
  evidence that utterance-processing entails mentalizing: {The} case of irony}.
\newblock \emph{NeuroImage}, 63(1):25--39.

\bibitem[{Stowe et~al.(2022)Stowe, Utama, and Gurevych}]{stowe_impli_2022}
Kevin Stowe, Prasetya Utama, and Iryna Gurevych. 2022.
\newblock \href {https://doi.org/10.18653/v1/2022.acl-long.369} {{IMPLI}:
  {Investigating} {NLI} {Models}' {Performance} on {Figurative} {Language}}.
\newblock In \emph{Proceedings of the 60th {Annual} {Meeting} of the
  {Association} for {Computational} {Linguistics} ({Volume} 1: {Long}
  {Papers})}, pages 5375--5388, Dublin, Ireland. Association for Computational
  Linguistics.

\bibitem[{Taylor et~al.(2022)Taylor, Kardas, Cucurull, Scialom, Hartshorn,
  Saravia, Poulton, Kerkez, and Stojnic}]{taylor_galactica_2022}
Ross Taylor, Marcin Kardas, Guillem Cucurull, Thomas Scialom, Anthony
  Hartshorn, Elvis Saravia, Andrew Poulton, Viktor Kerkez, and Robert Stojnic.
  2022.
\newblock \href {https://arxiv.org/abs/2211.09085} {Galactica: {A} {Large}
  {Language} {Model} for {Science}}.
\newblock {arXiv} preprint.

\bibitem[{Tessler and Franke(2018)}]{tessler_not_2018}
Michael~Henry Tessler and Michael Franke. 2018.
\newblock \href {https://cogsci.mindmodeling.org/2018/papers/0219/index.html}
  {Not unreasonable: {Carving} vague dimensions with contraries and
  contradictions}.
\newblock In \emph{Proceedings of the 40th {Annual} {Conference} of the
  {Cognitive} {Science} {Society}}.

\bibitem[{Tong et~al.(2021)Tong, Shutova, and Lewis}]{tong_recent_2021}
Xiaoyu Tong, Ekaterina Shutova, and Martha Lewis. 2021.
\newblock \href {https://doi.org/10.18653/v1/2021.naacl-main.372} {Recent
  advances in neural metaphor processing: {A} linguistic, cognitive and social
  perspective}.
\newblock In \emph{Proceedings of the 2021 {Conference} of the {North}
  {American} {Chapter} of the {Association} for {Computational} {Linguistics}:
  {Human} {Language} {Technologies}}, pages 4673--4686, Online. Association for
  Computational Linguistics.

\bibitem[{Trosborg(2010)}]{trosborg_pragmatics_2010}
Anna Trosborg, editor. 2010.
\newblock \href {https://doi.org/doi:10.1515/9783110214444} {\emph{Pragmatics
  across {Languages} and {Cultures}}}.
\newblock De Gruyter Mouton.

\bibitem[{Ullman(2023)}]{ullman_large_2023}
Tomer Ullman. 2023.
\newblock \href {https://arxiv.org/abs/2302.08399} {Large {Language} {Models}
  {Fail} on {Trivial} {Alterations} to {Theory}-of-{Mind} {Tasks}}.
\newblock {arXiv} preprint.

\bibitem[{Veatch(1998)}]{veatch_theory_1998}
Thomas~C. Veatch. 1998.
\newblock \href {https://doi.org/doi:10.1515/humr.1998.11.2.161} {A theory of
  humor}.
\newblock \emph{Humor}, 11(2):161--216.

\bibitem[{Vendetti et~al.(2019)Vendetti, Kamawar, and
  Andrews}]{vendetti_theory_2019}
Corrie Vendetti, Deepthi Kamawar, and Katherine~E. Andrews. 2019.
\newblock \href {https://doi.org/10.1037/dev0000666} {Theory of mind and
  preschoolers' understanding of misdeed and politeness lies}.
\newblock \emph{Developmental Psychology}, 55(4):823--834.

\bibitem[{Wang et~al.(2022)Wang, Mishra, Alipoormolabashi, Kordi, Mirzaei,
  Naik, Ashok, Dhanasekaran, Arunkumar, Stap, Pathak, Karamanolakis, Lai,
  Purohit, Mondal, Anderson, Kuznia, Doshi, Pal, Patel, Moradshahi, Parmar,
  Purohit, Varshney, Kaza, Verma, Puri, Karia, Doshi, Sampat, Mishra, Reddy~A,
  Patro, Dixit, and Shen}]{wang_super-naturalinstructions_2022}
Yizhong Wang, Swaroop Mishra, Pegah Alipoormolabashi, Yeganeh Kordi, Amirreza
  Mirzaei, Atharva Naik, Arjun Ashok, Arut~Selvan Dhanasekaran, Anjana
  Arunkumar, David Stap, Eshaan Pathak, Giannis Karamanolakis, Haizhi Lai,
  Ishan Purohit, Ishani Mondal, Jacob Anderson, Kirby Kuznia, Krima Doshi,
  Kuntal~Kumar Pal, Maitreya Patel, Mehrad Moradshahi, Mihir Parmar, Mirali
  Purohit, Neeraj Varshney, Phani~Rohitha Kaza, Pulkit Verma, Ravsehaj~Singh
  Puri, Rushang Karia, Savan Doshi, Shailaja~Keyur Sampat, Siddhartha Mishra,
  Sujan Reddy~A, Sumanta Patro, Tanay Dixit, and Xudong Shen. 2022.
\newblock \href {https://aclanthology.org/2022.emnlp-main.340}
  {Super-{NaturalInstructions}: {Generalization} via {Declarative}
  {Instructions} on 1600+ {NLP} {Tasks}}.
\newblock In \emph{Proceedings of the 2022 {Conference} on {Empirical}
  {Methods} in {Natural} {Language} {Processing}}, pages 5085--5109, Abu Dhabi,
  United Arab Emirates. Association for Computational Linguistics.

\bibitem[{Wei et~al.(2022)Wei, Bosma, Zhao, Guu, Yu, Lester, Du, Dai, and
  Le}]{wei_finetuned_2022}
Jason Wei, Maarten Bosma, Vincent Zhao, Kelvin Guu, Adams~Wei Yu, Brian Lester,
  Nan Du, Andrew~M. Dai, and Quoc~V. Le. 2022.
\newblock \href {https://openreview.net/forum?id=gEZrGCozdqR} {Finetuned
  {Language} {Models} are {Zero}-{Shot} {Learners}}.
\newblock In \emph{International {Conference} on {Learning} {Representations}}.

\bibitem[{Wilson and Sperber(2012)}]{wilson_meaning_2012}
D.~Wilson and D.~Sperber. 2012.
\newblock \href {https://books.google.com/books?id=wDTtW0L-P-MC} {\emph{Meaning
  and {Relevance}}}.
\newblock Cambridge University Press.

\bibitem[{Wilson and Sperber(1992)}]{wilson_verbal_1992}
Deirdre Wilson and Dan Sperber. 1992.
\newblock \href {https://doi.org/10.1016/0024-3841(92)90025-E} {On verbal
  irony}.
\newblock \emph{Lingua}, 87(1):53--76.

\bibitem[{Wolf et~al.(2020)Wolf, Debut, Sanh, Chaumond, Delangue, Moi, Cistac,
  Rault, Louf, Funtowicz, Davison, Shleifer, von Platen, Ma, Jernite, Plu, Xu,
  Le~Scao, Gugger, Drame, Lhoest, and Rush}]{wolf_transformers_2020}
Thomas Wolf, Lysandre Debut, Victor Sanh, Julien Chaumond, Clement Delangue,
  Anthony Moi, Pierric Cistac, Tim Rault, Remi Louf, Morgan Funtowicz, Joe
  Davison, Sam Shleifer, Patrick von Platen, Clara Ma, Yacine Jernite, Julien
  Plu, Canwen Xu, Teven Le~Scao, Sylvain Gugger, Mariama Drame, Quentin Lhoest,
  and Alexander Rush. 2020.
\newblock \href {https://doi.org/10.18653/v1/2020.emnlp-demos.6} {Transformers:
  {State}-of-the-{Art} {Natural} {Language} {Processing}}.
\newblock In \emph{Proceedings of the 2020 {Conference} on {Empirical}
  {Methods} in {Natural} {Language} {Processing}: {System} {Demonstrations}},
  pages 38--45, Online. Association for Computational Linguistics.

\bibitem[{Yoon et~al.(2016)Yoon, Tessler, Goodman, and
  Frank}]{yoon_talking_2016}
Erica~J. Yoon, Michael~Henry Tessler, Noah~D. Goodman, and Michael~C. Frank.
  2016.
\newblock \href {https://cogsci.mindmodeling.org/2016/papers/0477/index.html}
  {Talking with tact: {Polite} language as a balance between informativity and
  kindness}.
\newblock In \emph{Proceedings of the {Annual} {Meeting} of the {Cognitive}
  {Science} {Society}}.

\bibitem[{Yoon et~al.(2020)Yoon, Tessler, Goodman, and
  Frank}]{yoon_polite_2020}
Erica~J. Yoon, Michael~Henry Tessler, Noah~D. Goodman, and Michael~C. Frank.
  2020.
\newblock \href {https://doi.org/10.1162/opmi_a_00035} {Polite {Speech}
  {Emerges} {From} {Competing} {Social} {Goals}}.
\newblock \emph{Open Mind}, 4:71--87.

\bibitem[{Yule(1996)}]{yule_pragmatics_1996}
George Yule. 1996.
\newblock \emph{Pragmatics}, 1 edition.
\newblock Oxford {Introduction} to {Language} {Study}. Oxford University Press.

\bibitem[{Zadeh et~al.(2019)Zadeh, Chan, Liang, Tong, and
  Morency}]{zadeh_social-iq_2019}
Amir Zadeh, Michael Chan, Paul~Pu Liang, Edmund Tong, and Louis-Philippe
  Morency. 2019.
\newblock \href {https://doi.org/10.1109/CVPR.2019.00901} {Social-{IQ}: {A}
  {Question} {Answering} {Benchmark} for {Artificial} {Social} {Intelligence}}.
\newblock In \emph{2019 {IEEE}/{CVF} {Conference} on {Computer} {Vision} and
  {Pattern} {Recognition} ({CVPR})}, pages 8799--8809.

\bibitem[{Zheng et~al.(2021)Zheng, Qiu, Fan, Zhu, and Zhu}]{zheng_grice_2021}
Zilong Zheng, Shuwen Qiu, Lifeng Fan, Yixin Zhu, and Song-Chun Zhu. 2021.
\newblock \href {https://doi.org/10.18653/v1/2021.findings-acl.182} {{GRICE}:
  {A} {Grammar}-based {Dataset} for {Recovering} {Implicature} and
  {Conversational} {rEasoning}}.
\newblock In \emph{Findings of the {Association} for {Computational}
  {Linguistics}: {ACL}-{IJCNLP} 2021}, pages 2074--2085, Online. Association
  for Computational Linguistics.

\end{thebibliography}
\bibliographystyle{acl_natbib}

\appendix

\section{Example prompts}
\label{sec:example-prompts}

This section contains example prompts for each task in our experiments. See \Cref{sec:tasks} and \Cref{tab:examples} for details on the materials, and \Cref{sec:evaluation} for discussion of how prompts were constructed.

\setlength{\parindent}{0cm}

\subsection{Deceits}

\footnotesize
Task: You will read short stories that describe two characters interacting with each other. Each story will be followed by a multiple-choice question. Read each story and choose the best answer to each question. Your task is to decide why the character in the story responds in a certain way. The answer options are 1, 2, 3, or 4.
\linebreak

Scenario: Henry is sitting at his desk and watching TV, and reluctantly switches off the TV with the remote control and picks up a textbook. Shortly after, his mother comes in the room and asks, "What have you been doing up here?" Henry responds: "Reading." Why has Henry responded in such a way?

Options:

1) He has been reading for some time.

2) He does not want to offend his mom by not reading the books that she gave him.  

3) He does not want to get into trouble for not studying. 

4) He wants his mom to believe that he has been watching TV.

Answer: 

\subsection{IndirectSpeech}

\footnotesize
Task: You will read short stories that describe everyday situations. Each story will be followed by a multiple-choice question. Read each story and choose the best answer. Your task is to decide what the character in the story is trying to convey. The answer options are 1, 2, 3, or 4.
\linebreak

Scenario: Nate is about to leave the house. His wife points at a full bag of garbage and asks: "Are you going out?" What might she be trying to convey?

Options:

1) She wants Nate to spend more time with the family.

2) She wants to know Nate's plans.

3) She wants Nate to take the garbage out.

4) She wants Nate to bring his friends over. 

Answer:  

\subsection{Irony}

Task: You will read short stories that describe everyday situations. Each story will be followed by a multiple-choice question. Read each story and choose the best answer. Your task is to decide what the character in the story is trying to convey. The answer options are 1, 2, 3, or 4.
\linebreak

Scenario: It is a holiday. Stefan and Kim are sitting in the backseat of the car. They are fighting all the time. Their father says: "Oh, it is so pleasant here." What did the father want to convey?

Options:

1) He enjoys listening to his kids fighting.

2) He remembers about his wife's birthday.

3) He does not want to listen to his kids' arguments.

4) AC gives them some needed cool.

Answer: 

\subsection{Maxims}

\footnotesize
Task: You will read short stories that describe everyday situations. Each story will be followed by a multiple-choice question. Read each story and choose the best answer. Your task is to decide why the character in the story responds in a certain way. The answer options are 1, 2, 3, or 4.
\linebreak

Scenario: Leslie and Jane are chatting at a coffee shop. Leslie asks, "Who was that man that I saw you with last night?" Jane responds, "The latte is unbelievable here." Why has Jane responded like this?

Options:

1) She does not want to discuss the topic that Leslie has raised.

2) The man who Leslie saw makes unbelievable lattes.

3) She thinks that it is the best latte in the town.

4) A coffee break is not a good time to discuss men.

Answer: 

\subsection{Metaphor}

\footnotesize
Task: You will read short stories that describe everyday situations. Each story will be followed by a multiple-choice question. Read each story and choose the best answer to each question. The answer options are 1, 2, 3, 4, or 5.
\linebreak

Scenario: Andrew and Bob were discussing the investment company where Andrew works. Bob said: "The investors are squirrels collecting nuts." What does Bob mean?

Options:

1) The investors dress and eat well.

2) Squirrels were hired to work in the company. 

3) Bob is allergic to nuts.

4) They buy stocks hoping for future profit.  

5) The investors enjoy picking nuts as much as squirrels do.

Answer:  

\subsection{Humor}

\footnotesize
Task: You will read jokes that are missing their punch lines. A punch line is a funny line that finishes the joke. Each joke will be followed by five possible endings. Please choose the ending that makes the joke funny. The answer options are 1, 2, 3, 4, or 5.
\linebreak

Joke: Martha walked into a pastry shop. After surveying all the pastries, she decided on a chocolate pie. "I'll take that one," Martha said to the attendant, "the whole thing." "Shall I cut it into four or eight pieces?" the attendant asked. 

Punchlines:

1) Martha said, "My leg is hurting so much."

2) Martha said, "Four pieces, please; I'm on a diet."

3) Martha said: "Well, there are five people for dessert tonight, so eight pieces will be about right."

4) Then the attendant squirted whipped cream in Martha's face.

5) Martha said, "You make the most delicious sweet rolls in town."

Answer: 

\subsection{Coherence}

\footnotesize
Task: You will read pairs of sentences. Reach each pair and decide whether they form a coherent story. The answer options are 1 or 2.
\linebreak

Scenario: Cleo brushed against a table with a vase on it. She decided to study harder to catch up.

Options:

1) Incoherent

2) Coherent

Answer: 

\section{Timestamps of OpenAI model queries}
\label{sec:openai-timestamps}

\normalsize
\Cref{tab:openai_timestamps} shows timestamps of requests sent to the OpenAI API. 

\begin{table}[ht]
    \centering
    \scriptsize
    \begin{tabular}{ccc} \toprule
       Model & Phenomenon & Timestamp \\ \midrule
text-ada-001 & Coherence & 2022-10-11 12:28 -0400 \\
text-ada-001 & Deceits & 2022-10-11 12:28 -0400 \\
text-ada-001 & IndirectSpeech & 2022-10-11 12:28 -0400 \\
text-ada-001 & Irony & 2022-10-11 12:28 -0400 \\
text-ada-001 & Humor & 2022-10-11 12:28 -0400 \\
text-ada-001 & Maxims & 2022-10-11 12:29 -0400 \\
text-ada-001 & Metaphor & 2022-10-11 12:29 -0400 \\ \midrule
text-davinci-002 & Coherence & 2022-10-11 11:56 -0400 \\
text-davinci-002 & Deceits & 2022-10-11 11:55 -0400 \\
text-davinci-002 & IndirectSpeech & 2022-10-11 11:55 -0400 \\
text-davinci-002 & Irony & 2022-10-11 11:54 -0400 \\
text-davinci-002 & Humor & 2022-10-11 11:53 -0400 \\
text-davinci-002 & Maxims & 2022-10-11 11:56 -0400 \\
text-davinci-002 & Metaphor & 2022-10-11 11:57 -0400 \\ \bottomrule

    \end{tabular}
    \caption{Timestamps of OpenAI API model queries.}
    \label{tab:openai_timestamps}
\end{table}

\section{No-context analysis} \label{sec:no-context-analysis}

\subsection{Details of human experiments} \label{sec:human-expts}


\setlength{\parindent}{1.5em}


Below, we discuss details of the no-context human experiments described in \Cref{sec:role-of-context}. This study was approved by the Institutional Review Board at the home institution of the authors (protocol 2010000243).

\paragraph{Participants.} We collected data from 30 participants using Amazon.com's Mechanical Turk. All participants were recruited from IP addresses in the US, Canada, and other English-speaking countries and passed a brief English proficiency task to participate. We pre-screened participants using a qualification task in which they were asked to perform 10 simple sentence completions, which were judged for basic levels of coherence and grammaticality. Participants were paid 7 USD for completing the study, which took around 20 minutes to complete. The resulting hourly rate was around 21 USD, which is well above federal minimum wage in the United States. 

\paragraph{Procedure.} Participants completed these tests during one individual testing session. After giving informed consent, which included assurance of anonymity, participants were shown instructions and a training trial, in which they were told they would be answering questions about a character in a short interaction. They then saw 105 trials (similar to those described in \Cref{sec:example-prompts}), without the scenario context. For example:\newline

\footnotesize

\setlength{\parindent}{0em}

Bob said: "The investors are squirrels collecting nuts." What does Bob mean?

1) The investors dress and eat well.

2) Squirrels were hired to work in the company. 

3) Bob is allergic to nuts.

4) They buy stocks hoping for future profit.  

5) The investors enjoy picking nuts as much as squirrels do. \newline

\normalsize

Items were presented within blocks according to their phenomenon, as in \citeauthor{floyd_deciphering_nodate}'s (\citeyear{floyd_deciphering_nodate}) original experiments. Blocks and items were presented in a random order.

\begin{figure}[t]
    \centering
    \includegraphics[width=\linewidth]{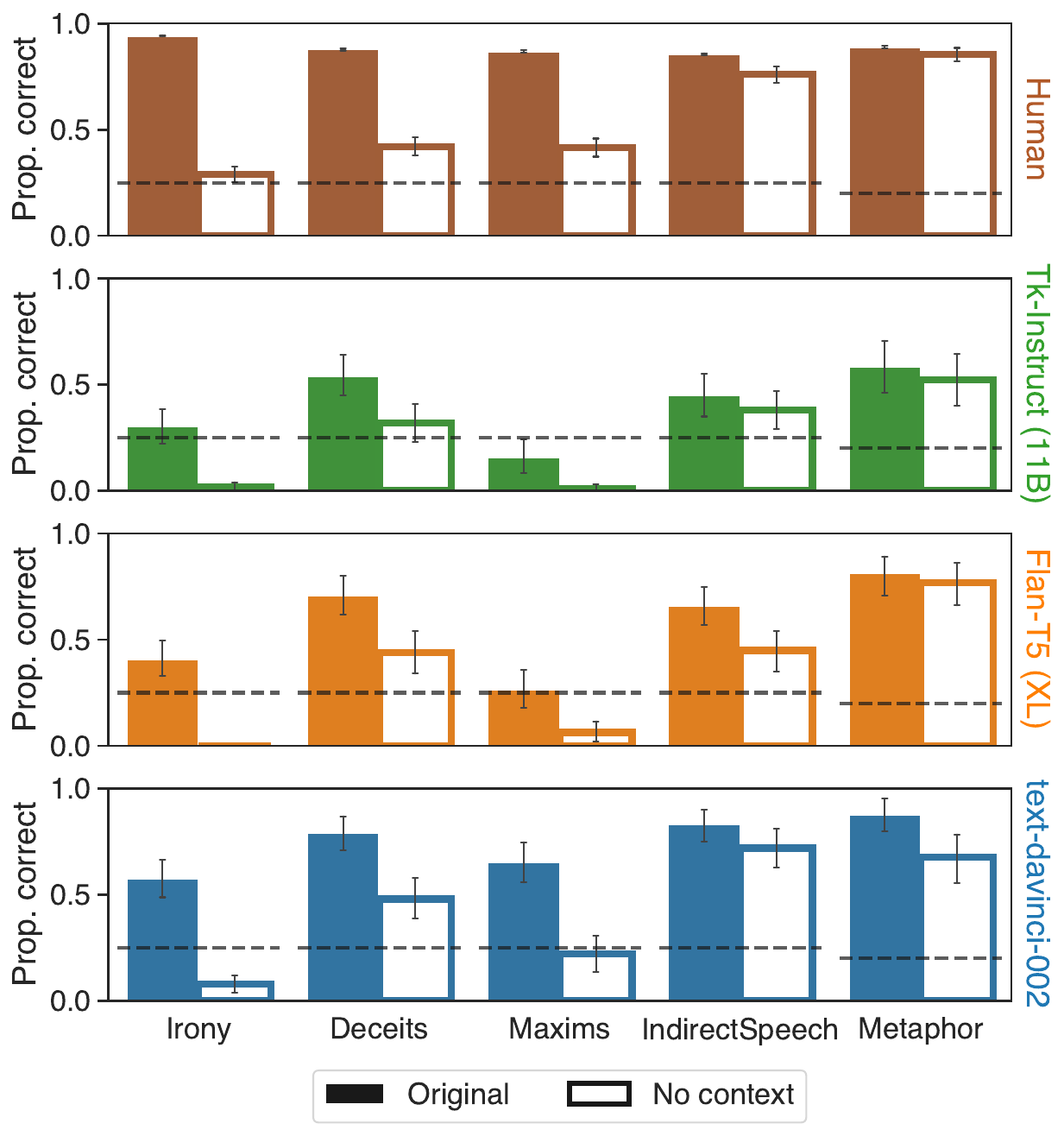}
    \caption{Proportion of items where humans and models select the correct pragmatic answer, on both original (shaded bars) and no-context (empty bars) versions.}
    \label{fig:no-context-raw}
\end{figure}

\begin{figure*}[ht]
    \centering
    \includegraphics[width=\linewidth]{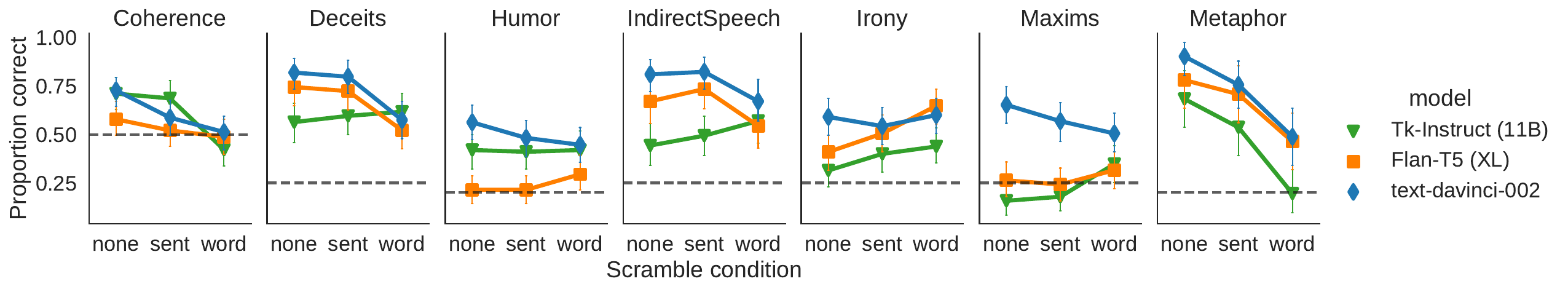}
    \caption{Model performance across scrambling conditions (none $=$ original, unmodified items). Error bars denote 95\% CI. Dashed line indicates random baseline.}
    \label{fig:scrambling}
\end{figure*}

\subsection{Raw accuracy scores} \label{sec:no-context-raw}

\Cref{fig:no-context-raw} shows accuracy scores achieved by humans and the three best-performing models on the original (shaded bars) and no-context (empty bars) versions of the test items.

\section{Sentence- and word-level scrambling} \label{sec:scrambling}

\Cref{fig:scrambling} shows accuracy scores achieved by the three best-performing models on each task, across three scrambling conditions: none (original, unmodified items), sentence-level, and word-level. Example prompts are provided below.

\subsection{Sentence-level scrambled prompt}

\footnotesize
Task: You will read short stories that describe two characters interacting with each other. Each story will be followed by a multiple-choice question. Read each story and choose the best answer to each question. Your task is to decide why the character in the story responds in a certain way. The answer options are 1, 2, 3, or 4.
\linebreak

Scenario: Dan says,"The dog knocked it over." The vase falls down on the floor and breaks. He brushes against his mother's vase. When Dan's mother comes home, she asks Dan: "What happened to my vase?" Dan is playing in the living room. Why has Dan responded in such a way? 

Options: 

1) Dan does not want his mom to be angry with him for breaking the vase. 

2) Dan finds this vase ugly and wants to get rid of it. 

3) Dan wants his mom to know that he knocked it over. 

4) Dan thinks that the dog has knocked over the vase. 

Answer: 

\subsection{Word-level scrambled prompt}

Task: You will read short stories that describe two characters interacting with each other. Each story will be followed by a multiple-choice question. Read each story and choose the best answer to each question. Your task is to decide why the character in the story responds in a certain way. The answer options are 1, 2, 3, or 4. 
\linebreak

Scenario: to happened Dan "The against in it she comes "What living Dan the vase floor on down The Dan: He dog my brushes vase?" mother When falls breaks. vase. and playing room. his asks knocked says, home, over." the mother's is Dan's Why has Dan responded in such a way? 

Options: 

1) Dan does not want his mom to be angry with him for breaking the vase. 

2) Dan finds this vase ugly and wants to get rid of it. 

3) Dan wants his mom to know that he knocked it over. 

4) Dan thinks that the dog has knocked over the vase. 

Answer: 

\end{document}